\documentclass[11pt,twoside]{article}

\usepackage{jair}
\usepackage{theapa}
\usepackage{rawfonts}

\jairheading{78}{2023}{781-821}{05/2023}{11/2023}
\ShortHeadings{Learning Criteria Going Beyond the Usual Risk}
{Holland \& Tanabe}
\firstpageno{781}

\usepackage{./config/all}

\usepackage{natbib} 
\usepackage{amsmath} 
\usepackage{amssymb} 
\usepackage{amsfonts} 
\usepackage{bm} 
\usepackage{amsthm} 
\theoremstyle{definition} \newtheorem{defn}{Definition}
\theoremstyle{plain} 
\theoremstyle{plain} 
\theoremstyle{plain} 
\theoremstyle{plain} 
\theoremstyle{remark} \newtheorem{rmk}[defn]{Remark}
\theoremstyle{remark} \newtheorem{ex}[defn]{Example}

\newcommand{\term}[1]{\textcolor{BlueViolet}{\textit{{#1}}}}

\usepackage{enumitem}
\makeatletter
\def\namedlabel#1#2{\begingroup
    #2%
    \def\@currentlabel{#2}%
    \phantomsection\label{#1}\endgroup
}
\makeatother

\newcommand*{\defeq}{\mathrel{\vcenter{\baselineskip0.5ex \lineskiplimit0pt     
                     \hbox{\scriptsize.}\hbox{\scriptsize.}}}=}

\newcommand{\overbar}[1]{\mkern 1.5mu\overline{\mkern-1.5mu#1\mkern-1.5mu}\mkern 1.5mu}

\newcommand*\dif{\mathop{}\!\mathrm{d}}

\DeclareMathOperator*{\argmin}{arg\,min}
\DeclareMathOperator*{\argmax}{arg\,max}

\newcommand{\Abs}[1]{\lVert{#1}\rVert} 
\newcommand{\abs}[1]{\lvert{#1}\rvert} 
\DeclareMathOperator{\atan}{atan} 
\def\bigO{\mathcal{O}} 
\def\CC{\mathcal{C}} 
\DeclareMathOperator{\cpt}{CPT} 
\DeclareMathOperator{\crit}{C} 
\def\cond{\,\vert\,} 
\def\ddist{\mu} 
\DeclareMathOperator{\dfun}{F} 
\DeclareMathOperator{\Div}{Div} 
\DeclareMathOperator{\exx}{\mathbf{E}} 
\def\HH{\mathcal{H}} 
\DeclareMathOperator{\indic}{I} 
\def\LL{\mathcal{L}} 
\DeclareMathOperator{\loss}{\rdv{L}} 
\DeclareMathOperator{\myw}{w} 
\def\NN{\mathbb{N}} 
\def\PP{\mathcal{P}} 
\DeclareMathOperator{\prr}{\mathbf{P}} 
\DeclareMathOperator{\quant}{Q} 
\newcommand{\rdv}[1]{\mathsf{#1}} 
\def\RR{\mathbb{R}} 
\DeclareMathOperator{\sign}{sign} 
\DeclareMathOperator{\vaa}{var} 
\def\valfn{V} 
\def\XX{\mathcal{X}} 
\def\Z{\mathbf{Z}} 

\usepackage[pdfauthor={Matthew J. Holland},%
colorlinks=true,
linkcolor=blue,%
citecolor=blue]{hyperref}
\hypersetup{pdftitle={A Survey of Learning Criteria Going Beyond the Usual Risk}} 

\usepackage{etoolbox}
\makeatletter
\patchcmd{\NAT@citex}
  {\@citea\NAT@hyper@{%
     \NAT@nmfmt{\NAT@nm}%
     \hyper@natlinkbreak{\NAT@aysep\NAT@spacechar}{\@citeb\@extra@b@citeb}%
     \NAT@date}}
  {\@citea\NAT@nmfmt{\NAT@nm}%
   \NAT@aysep\NAT@spacechar\NAT@hyper@{\NAT@date}}{}{}

\patchcmd{\NAT@citex}
  {\@citea\NAT@hyper@{%
     \NAT@nmfmt{\NAT@nm}%
     \hyper@natlinkbreak{\NAT@spacechar\NAT@@open\if*#1*\else#1\NAT@spacechar\fi}%
       {\@citeb\@extra@b@citeb}%
     \NAT@date}}
  {\@citea\NAT@nmfmt{\NAT@nm}%
   \NAT@spacechar\NAT@@open\if*#1*\else#1\NAT@spacechar\fi\NAT@hyper@{\NAT@date}}
  {}{}
\makeatother

\begin{document}

\title{A Survey of Learning Criteria Going Beyond the Usual Risk}

\author{\name Matthew J.~Holland \email matthew-h@ar.sanken.osaka-u.ac.jp \\
       \name Kazuki Tanabe \email tanabe@ar.sanken.osaka-u.ac.jp \\
       \addr Osaka University\\
       Ibaraki, Osaka 567-0047 Japan}

\maketitle

\begin{abstract}
Virtually all machine learning tasks are characterized using some form of loss function, and ``good performance'' is typically stated in terms of a sufficiently small \emph{average} loss, taken over the random draw of test data. While optimizing for performance on average is intuitive, convenient to analyze in theory, and easy to implement in practice, such a choice brings about trade-offs. In this work, we survey and introduce a wide variety of non-traditional criteria used to design and evaluate machine learning algorithms, place the classical paradigm within the proper historical context, and propose a view of learning problems which emphasizes the question of ``what makes for a desirable loss distribution?'' in place of tacit use of the expected loss.
\end{abstract}



\section{Introduction}\label{sec:intro}

Over the past two decades, the rapid and widespread adoption of machine learning tools has provided a major impetus to design new methods and develop a more refined methodology. It is only natural that as machine learning techniques are brought to bear on new and increasingly diverse problems, it becomes necessary to re-define what we mean by ``good'' performance. Put roughly, within the classical paradigm of statistical machine learning, ``good'' invariably means ``small average test loss.'' While this is a perfectly natural definition, and in some cases precisely what we want to achieve, prioritizing the average test loss means that a lot of other important aspects of performance are abstracted away, and left without any guarantee. As a result, even when a solution is optimal on average, the results may be highly unsatisfactory. Here are a few examples to emphasize this point:
\begin{itemize}
\item \textbf{High-confidence guarantees.} Assuming just finite variance, in terms of the \emph{average} squared estimation error of the mean, the sample mean is minimax optimal \citep{bickel1981,lugosi2019b}. However, in terms of \emph{high-probability} error bounds, the sample mean has exponentially worse dependence on the confidence level than more robust alternatives \citep{catoni2012a,devroye2016a}.

\item \textbf{Balanced classification.} Under extreme class imbalance, even the best possibile misclassification probability can result in very poor performance on minority classes, leading to a large \term{balanced error} \citep{menon2021a}.

\item \textbf{Fair decisions.} Even if a dataset does not include sensitive demographic information such as ethnicity, gender, or economic status, optimal performance on average can lead to highly sub-optimal performance for under-represented groups \citep{hashimoto2018a}.

\item \textbf{Distribution shift.} Superior performance on average during training could be disastrous at test time if the underlying data distribution has changed \citep{curi2020a,quinonero2008DataShiftML,sutter2021a}.

\item \textbf{Resilience to rare events.} Decisions which are optimal on average need not be resilient to rare but catastrophic ``black swan'' events \citep{lee2020a,taleb2008BlackSwan}.
\end{itemize}
These issues highlight the fact that a successful machine learning application need not always be characterized by achieving the minimum average loss. This broadened perspective has driven the machine learning community to develop a wide variety of new techniques to ensure learning algorithms reliably provide desirable solutions under more diverse notions of success. In particular, over the past five years, clear trends in algorithm design have emerged in which the underlying loss function is left abstract, and \emph{properties of the loss distribution} are the primary focus. Our chief goal in this paper is to elucidate and bring attention to this trend. We survey the recent literature, provide some historical context, and use a general formulation that allows us to concisely express a wide variety of learning problems as special cases.

\subsection{The Classical Paradigm}\label{sec:intro_classical}

To make it clear precisely what this paper is about, we start with a concise formulation of the traditional learning problem. In the classical paradigm, one seeks the best possible performance \emph{on average}. Put simply, this refers to the fact that in the vast majority of statistical learning problems, the ultimate goal is formulated in terms of minimizing the expected ``test'' loss. More formally, letting $\rdv{Z} \sim \ddist$ denote an as-yet unobserved random (test) data point, $\HH$ a set of possible decisions, and $\ell(\cdot)$ a loss function which assigns a numerical penalty to $(\text{decision},\text{data})$ pairs, the ultimate goal is to find a solution $h^{\ast}$, which may or may not belong to $\HH$, that satisfies
\begin{align}\label{eqn:classic_risk_min}
\exx_{\ddist}\ell(h^{\ast};\rdv{Z}) \leq \inf_{h \in \HH} \exx_{\ddist}\ell(h;\rdv{Z}) + \varepsilon
\end{align}
for some $\varepsilon \geq 0$, which represents our tolerance level. The expected value of a random loss is traditionally called the \term{risk}, a term with a long history (see \S{\ref{sec:history}} for details). Depending on the actual realization of $\rdv{Z}$, the quality of any $h$ as measured by $\ell(h;\rdv{Z})$ is subject to variability. While performance could certainly be catastrophic in the worst case, the milder condition (\ref{eqn:classic_risk_min}) simply requires that on average, the decision made is optimal.\footnote{It is sometimes meaningless to ask for worst-case optimality over the random draw of test data. For example, consider estimating the mean of $\rdv{X} \sim \text{Normal}(0,1)$. For any choice of $\theta \in \RR$, we always have $\sup_{\rdv{X}}(\theta - \rdv{X})^{2} = \infty$.} Even so, the problem is still far from trivial, since $\ddist$ is unknown. This is where ``training'' data comes in. Let us denote a sequence of random training data by $\Z_{n} \defeq (\rdv{Z}_{1},\ldots,\rdv{Z}_{n})$. Obviously, if there is no relation between the distributions of $\Z_{n}$ and $\rdv{Z}$, we have no hope of finding a solution satisfying (\ref{eqn:classic_risk_min}). A strong relation can be made by the following assumption:
\begin{align}\label{eqn:iid_training_data}
\Z_{n} \text{ is an independent random sample of } n \text{ points drawn from } \ddist.
\end{align}
Seeking a solution to (\ref{eqn:classic_risk_min}) under the assumption (\ref{eqn:iid_training_data}) is precisely what \citet[Ch.~1]{vapnik1999NSLT} calls the ``general setting of the learning problem.'' With this link between training and test data in place, the most natural strategy is to replace the unknown $\ddist$ in (\ref{eqn:classic_risk_min}) with the empirical distribution induced by $\Z_{n}$, and optimize that objective instead; this is \term{empirical risk minimization (ERM)}. More formally, an ERM solution is any decision $\rdv{H}_{n}$ satisfying the following property:
\begin{align}\label{eqn:classic_erm_condition}
\rdv{H}_{n} \in \argmin_{h \in \HH} \frac{1}{n} \sum_{i=1}^{n} \ell(h;\rdv{Z}_{i}).
\end{align}
Intuitively, the hope is that a solution which is optimal at training time will also be optimal at test time. Of course, this is an idealized goal, since with limited samples, true optimality is not statistically plausible. Furthermore, the off-sample quality depends entirely on the nature of the random training data that happens to materialize. It is always possible to get a very unlucky sample $\Z_{n}$ which is not at all representative of $\ddist$ and leads to a poor candidate $\rdv{H}_{n}$, but under the assumption (\ref{eqn:iid_training_data}), this cannot occur with high probability. With this in mind, it is common to relax the requirement (\ref{eqn:classic_risk_min}), and ask that for some $\varepsilon > 0$ we have
\begin{align}\label{eqn:classic_pac_condition}
\prr\left\{ \exx_{\ddist}\ell(\rdv{H}_{n};\rdv{Z}) \leq \inf_{h \in \HH} \exx_{\ddist}\ell(h;\rdv{Z}) + \varepsilon \right\} \geq 1 - \delta
\end{align}
with $0 < \delta < 1$ controlling the desired level of confidence, noting that probability is with respect to $\Z_{n}$ and any algorithmic randomness that choosing $\rdv{H}_{n}$ entails.\footnote{When $\rdv{H}_{n} \notin \HH$ is allowed under (\ref{eqn:classic_pac_condition}), the learning problem is sometimes called \term{improper} \citep[Ch.~3]{shalev2014a}.} A great deal of statistical learning theory amounts to finding a bound on the sample size $n$ that is sufficient or necessary for $\rdv{H}_{n}$, the output of a $\Z_{n}$-dependent learning algorithm, to satisfy the performance requirement (\ref{eqn:classic_pac_condition}) over various classes of data distributions $\ddist$, decision spaces $\HH$, and loss functions $\ell$ \citep{anthony1999NNTheory,kearns1994CLTIntro,mohri2012Foundations,vapnik1998SLT,vapnik1999NSLT}.

\subsection{Our Viewpoint}\label{sec:intro_our_view}

On one hand, the breadth of learning problems captured within the classical paradigm of \S{\ref{sec:intro_classical}} is tremendous. Even with $\HH$ fixed, different tasks can be characterized by distinct loss functions, and a massive body of knowledge regarding effective loss function design has been developed in the statistical machine learning community, with countless refinements and extensions made to suit the needs of applied fields such as computer vision and natural language processing.

On the other hand, the class of problems captured within \S{\ref{sec:intro_classical}} can also be seen as being severely limited by the (often tacit) assumption that $\exx_{\ddist}\ell(\cdot;\rdv{Z})$ is the appropriate criterion to use for a given loss function. Taking a broader viewpoint, the expected value $\exx_{\ddist}\ell(h;\rdv{Z})$ is just one ``property'' of the distribution of the random loss $\ell(h;\rdv{Z})$ incurred by decision candidate $h \in \HH$. As the machine learning methodology develops and matures, we believe that the question of \emph{which property to emphasize as the criterion for learning} will play a critical role, in some cases orthogonal to the choice of $\HH$ and $\ell$.

Making this all a bit more explicit, we take $\ell$ and $\HH$ as given, and focus on the random losses under the test data distribution $\ddist$, denoted by
\begin{align}\label{eqn:loss_distro_set}
\LL_{\ddist}(\HH) \defeq \{ \ell(h;\rdv{Z}): \rdv{Z} \sim \ddist, h \in \HH \}.
\end{align}
Note also that the learning problem described in (\ref{eqn:classic_risk_min}) can be equivalently re-stated in terms of \emph{choosing a desirable loss distribution}, that is, seeking $\loss_{\ast}$ such that
\begin{align}
\exx_{\ddist}\loss_{\ast} \leq \exx_{\ddist}\loss + \varepsilon, \quad \forall \, \loss \in \LL_{\ddist}(\HH).
\end{align}
Here ``desirability'' is measured by reducing a random variable to a one-number summary, i.e., the mapping $\loss \mapsto \exx_{\ddist}\loss \in \overbar{\RR}$. If learning is broadly understood as an attempt at decision-making which leads to a desirable off-sample loss distribution, then aside from technical convenience and historical precedent, there is in principle no reason to restrict ourselves to the mean. For example, any valid median, the variance, skewness, kurtosis, inter-quartile range, weighted sums of these quantities; these are all elementary examples of properties which might be considered when making a broadened notion of ``desirable loss distribution'' precise.

In this paper, we generalize the class of learning problems described in \S{\ref{sec:intro_classical}} as follows. Let $\crit_{\ddist}:\LL_{\ddist}(\HH) \to \overbar{\RR}$ be any function which maps random variables $\loss \in \LL_{\ddist}(\HH)$ to the extended real line, and for readability let us overload the notation setting $\crit_{\ddist}(h) \defeq \crit_{\ddist}(\ell(h;\rdv{Z}))$ for each $h \in \HH$. We call $\crit_{\ddist}(\cdot)$ a \term{learning criterion} function, and refer to the value $\crit_{\ddist}(\loss)$ as a \term{property} of the distribution of $\loss$. Finally, fix a subset $\CC \subset \overbar{\RR}$ that we call the \term{desirability set}. With these parts in place, we expand upon the traditional learning problem of (\ref{eqn:classic_pac_condition}) using the following general condition on algorithm output $\rdv{H}_{n}$:
\begin{align}\label{eqn:new_pac_condition}
\prr\left\{ \crit_{\ddist}(\rdv{H}_{n}) \in \CC \right\} \geq 1 - \delta.
\end{align}
In words, this generalized setup can be succinctly summarized as follows.
\begin{center}
\underline{\textbf{Generalized learning problem:}}\\
\smallskip
``$\underbrace{\text{Reliably}}_{\geq 1-\delta}$ achieve a $\underbrace{\text{test loss}}_{\loss \in \LL_{\ddist}(\HH)}$ distribution with $\underbrace{\text{sufficiently desirable}}_{\CC}$ $\underbrace{\text{properties}}_{\crit_{\ddist}}$.''
\end{center}
Moving forward, we shall use this general formulation as a vantage point from which we can view recent conceptual and technical trends in machine learning as special cases. Our focus in this paper is purely on the criterion $\crit_{\ddist}(\cdot)$. Issues of loss function design, regularization, and model properties/constraints are all abstracted away, implicitly captured by $\ell$, $\HH$, and $\CC$. For reference, however, here are some simple and concrete examples illustrating the flexibility of the formulation given here.

\begin{ex}[Supervised learning]
Let $\rdv{Z} = (\rdv{X},\rdv{Y})$, with $\rdv{X}$ a random feature vector, and $\rdv{Y}$ a random binary label taking values in $\{-1,1\}$ for example. Given a classifier $h$, the (random) hinge loss incurred by $h$ is $\max\{0,1-\rdv{Y}h(\rdv{X})\}$, and the zero-one loss is $\indic\{h(\rdv{X}) \neq \rdv{Y}\}$.\hfill$\blacksquare$
\end{ex}

\begin{ex}[Unsupervised learning]
Let $\rdv{X}$ and $h(\rdv{X})$ be random vectors taking values on a linear space equipped with a norm $\Abs{\cdot}$, where $h(\rdv{X})$ satisfies some special constraints, e.g., a neural network whose intermediate layers provide a low-memory encoding of $\rdv{X}$, and use the distortion $\Abs{h(\rdv{X})-\rdv{X}}$ as a loss.\hfill$\blacksquare$
\end{ex}

\begin{ex}[Standard setup]\label{ex:intro_standard}
By setting the learning criterion and desirability set as
\begin{align*}
\crit_{\ddist}(\loss) = \exx_{\ddist}\loss, \quad \CC = (-\infty, \crit_{\ddist}^{\ast}+\varepsilon]
\end{align*}
we get a familiar formulation of the learning problem, with some flexibility left in specifying $\CC$, depending on how we set the threshold $\crit_{\ddist}^{\ast}$. If we set
\begin{align*}
\crit_{\ddist}^{\ast} = \inf\left\{\crit_{\ddist}(h): h \in \HH\right\}
\end{align*}
then we recover (\ref{eqn:classic_pac_condition}) as a special case of (\ref{eqn:new_pac_condition}). If instead constraining ourselves to $\HH$, we set
\begin{align*}
\crit_{\ddist}^{\ast} = \inf\left\{\crit_{\ddist}(h): h \text{ satisfies } \abs{\crit_{\ddist}(h)} < \infty\right\}
\end{align*}
then the desirability set $\CC$ encodes a requirement to achieve a small \term{approximation error} on top of a small \term{estimation error}, following the terminology of \citet{cucker2002a}.\hfill$\blacksquare$
\end{ex}

\begin{ex}[A less typical setup]\label{ex:intro_deviations}
One can in principle consider setting a threshold $\crit_{\ddist}^{\ast}$ about which we hope performance does not deviate too far, namely
\begin{align*}
\CC = \left[ \crit_{\ddist}^{\ast}-\underline{\varepsilon}, \crit_{\ddist}^{\ast}+\overline{\varepsilon} \right], \quad \overline{\varepsilon}, \underline{\varepsilon} \geq 0
\end{align*}
where the width $\overline{\varepsilon} + \underline{\varepsilon}$ controls the amount of deviation that we tolerate. Note that while modifying the design of $\crit_{\ddist}$ lets us encode preferences on the deviations of $\loss$, introducing $\CC$ in this form enables us to request small deviations over the random draw of $\Z_{n}$ as well.\hfill$\blacksquare$
\end{ex}

\begin{ex}[Relative performance]\label{ex:intro_relative}
Another interesting way to define desirability through $\crit_{\ddist}(\cdot)$ and a threshold $\crit_{\ddist}^{\ast}$ is to look at relative values. For example, letting $a,b,c$ be non-negative real values, consider the desirability set
\begin{align*}
\CC = \left\{ x \in \RR: \frac{\abs{x - \crit_{\ddist}^{\ast}}}{a + b\abs{x} + c\abs{\crit_{\ddist}^{\ast}}} \leq \varepsilon \right\}
\end{align*}
where $\varepsilon > 0$ once again specifies a tolerance level. See \citet{haussler1992a} for an example of learning theoretical results stated in terms of this notion of desirability, with $a > 0$ and $b=c=1$.\hfill$\blacksquare$
\end{ex}

\subsection{Related Work}\label{sec:intro_related_work}

Now that we have a picture of what this paper is about, let us first take a bird's-eye view of the context in which this survey was born, and highlight some topics that we do \emph{not} cover.

Starting with some clear omissions, we do not delve into reinforcement learning at all, although the basic viewpoint put forward in \S{\ref{sec:intro_our_view}} can clearly be extended to such problem settings; see \citet{chow2016a,chow2017a} for some examples. We also do not explicitly treat any Bayesian learning theory, but a great deal of such theory can indeed be captured by our formulation when $\HH$ is a set of sufficiently regular probability distributions over a parameter space, and $\ell$ or $\HH$ are designed to enforce some degree of ``proximity'' to a prior distribution. To see this, consider for example the rich examples derived from the general ``Bayesian Learning Rule'' of \citet{khan2021arxiv}. We do not delve into any details regarding the duality theory underlying convex learning criteria; see the very recent survey of \citet{royset2022arxiv} for more detailed background. We also do not provide a comprehensive list of applied research featuring machine learning driven by novel learning criteria; the recent survey of \citet{hu2022a} provides a useful reference in this respect. Finally, since the underlying base loss function is left abstract, we do not consider adversarial methods which depend on particular forms of losses; see recent work by \citet{robey2022a} which interpolates between average and worst-case criteria by controlling perturbations of the inputs used in supervised learning tasks.

We used the broad term ``property'' to refer to the output of criterion $\crit_{\ddist}$ in \S{\ref{sec:intro_our_view}}, and while we believe this general notion is not yet widespread in the machine learning community, the idea of defining various characteristics of interest for an unknown (non-parametric) distribution is certainly very well-established in statistics.\footnote{Work by Lehmann, Bickel, and Huber in the 1960s and 1970s was particularly influential in this respect. See the commentary of \citet{oja2012} for some illuminating historical context.} It should also be noted that the term \term{property} has been used in a formal way, and that there is a broad literature on the \term{elicitation} of properties, i.e., characterizing properties which can be expressed as the solution to an expected loss minimization problem. An important subset of such properties are the ``M-parameters'' as described lucidly by \citet{koltchinskii1997a}, an extension of classical M-estimators \citep{huber1964a}. The literature on property elicitation begins in statistics and economics \citep{lambert2008a,savage1971a}, but important connections to surrogate complexity in machine learning are being established \citep{finocchiaro2021a}. Elicitation and surrogate function design are deep and interesting topics, especially in the generalized learning setup of \S{\ref{sec:intro_our_view}}, but in this paper, we will not delve further in this direction. 

As emphasized in the previous sub-section, the central idea here is that of putting the loss distribution at the forefront, making ``desirable properties'' a key design decision, rather than falling back on the tacit assumption that a small mean is always good enough. While up until recently it has remained rather obscure, this notion has existed in the machine learning community for decades. For a lucid example in the context of binary classification, \citet{breiman1999a} gave an example showing how a uniformly higher margin distribution need not imply better off-sample classification performance. \citet{reyzin2006a} built upon this example, showing how naive margin maximization tends to be inferior to algorithms which ``do a better job with the overall distribution.'' Two decades later, the idea of taking a broader view of loss distributions is starting to receive widespread attention, in terms of both designing and evaluating algorithms. One stark case is the work of \citet{agarwal2021a}, who used tail-trimmed confidence intervals to effectively infer about deep reinforcement learning algorithm performance after just ``a handful of runs,'' receiving an Outstanding Paper Award at NeurIPS 2021.

The learning criterion $\crit_{\ddist}$ is something to be explicitly \emph{designed}, and coupled with desirability conditions encoded by $\CC$, it ultimately encodes an \emph{optimization} task to be solved by the machine. In this direction, the massive literature on \term{risk functions} (also \term{risk measures}) in financial settings is of great relevance. Letting $\loss \in \LL_{\ddist}(\HH)$ represent an uncertain financial outcome (e.g., negative returns of a certain portfolio), a succinct numerical summary of this uncertainty via $\loss \mapsto \crit_{\ddist}(\loss)$ is important for guiding principled decision-making under uncertainty. The notion of \term{coherent} risk functions as axiomatized by \citet{artzner1999a} has been extremely influential, and important special cases such as \term{conditional value-at-risk (CVaR)} have been central to both theory and practice of financial risk design. Sufficiently regular risk functions (e.g., coherent risks) can be used to characterize \term{deviation} properties \citep{rockafellar2006a}, and the ``risk quadrangle'' of \citet{rockafellar2013a} gives a detailed systematic account of functions for quantifying risk and deviations, linking them up with notions of ``regret'' and ``error.'' In a sense, the basic viewpoint and degree of generality in our formulation from \S{\ref{sec:intro_our_view}} is also present in \citep{rockafellar2013a}, as is evident in the following quote:
\begin{quote}
A decision [results in] \emph{random variables}\ldots which can only be \emph{shaped in their distributions} through the [decision made], not pinned down to specific values. Now there is no longer a single, evident answer to how optimization should be viewed, but risk measures can come to the rescue.\\
\smallskip
\hfill---\enskip\citet[\S{5}]{rockafellar2013a}
\end{quote}
Traditionally, decisions are made so as to minimize risk, however it may be defined. Our viewpoint differs slightly in that all four ``corners'' of the risk quadrangle are homogenized and placed on equal footing as potential candidates for the learning criterion $\crit_{\ddist}$.

Our general formulation of $\loss$ being ``sufficiently desirable'' in terms of $\crit_{\ddist}(\loss) \in \CC$ has been influenced by the well-known learning model of \citet{haussler1992a}. In this model, $\crit_{\ddist}(\loss) = \exx_{\ddist}\loss$ is assumed throughout (called the ``risk''), but another notion called ``regret'' is introduced, a non-negative value which evaluates the desirability of $\loss$ as measured by $\exx_{\ddist}\loss$. Using Haussler's definition, the (random) regret in the formulation (\ref{eqn:classic_pac_condition}) is the Bernoulli random variable 
\begin{align*}
\indic\left\{ \exx_{\ddist}\ell(\rdv{H}_{n};\rdv{Z}) \leq \inf_{h \in \HH} \exx_{\ddist}\ell(h;\rdv{Z}) + \varepsilon \right\},
\end{align*}
whereas in our generalized condition (\ref{eqn:new_pac_condition}), it is
\begin{align*}
\indic\left\{ \crit_{\ddist}(\rdv{H}_{n}) \in \CC \right\}.
\end{align*}
Since Haussler's full learning problem requires us to choose $\rdv{H}_{n}$ to minimize the expected regret over the draw of $\Z_{n}$, our setup in \S{\ref{sec:intro_our_view}} is a special case of the framework in \citep{haussler1992a} whenever we restrict ourselves to $\crit_{\ddist}(\loss) = \exx_{\ddist}\loss$. On the other hand, with the freedom to design $\crit_{\ddist}(\cdot)$ as we like, we are able to go well beyond the original Haussler model.

\subsection{Clerical Matters}

We use \emph{plain italics} to indicate emphasis, and \term{colored italics} to indicate terms that have formal definitions, either within this paper or in the works we have cited. We use ``$\blacksquare$'' to indicate the end of examples and remarks. We distinguish random quantities by using sans-serif typeface, e.g., $\rdv{X}$, $\rdv{Z}$, $\rdv{Z}_{i}$, $\rdv{H}_{n}$, $\widehat{\rdv{C}}_{n}$, and so forth. Throughout this paper, $\ddist$ denotes the probability distribution of the underlying random data of interest; sometimes this will be $\rdv{X}$, sometimes this will be $(\rdv{X},\rdv{Y})$, sometimes this will be an abstract $\rdv{Z}$. In any case, expectation will be denoted by $\exx_{\ddist}$. Probabilities on the other hand are expressed using a generic symbol $\prr$, regardless of the underlying distribution. The event ``random variable $\rdv{X}$ equals $x$'' is written $\{\rdv{X} = x\}$, and the corresponding Bernoulli random variable (via indicator function) is $\indic\{\rdv{X} = x\}$. For any positive integer $k$, we write $[k] \defeq \{1,\ldots,k\}$, useful for indexing.

\section{Core Concepts}\label{sec:core_concepts}

From a historical perspective, there have always been clear motivations, both theoretical and practical, for introducing new learning criteria into the picture. In this section, we briefly highlight some of the key scenarios within which these functions have arisen. A more systematic and general exposition of important criterion classes will follow in \S{\ref{sec:lcrit_classes}}.

\subsection{Robustness Under Heavy-Tailed Data}\label{sec:core_concepts_robustness}

As mentioned in \S{\ref{sec:intro}}, one of the most natural reasons to design novel objective functions is that the empirical mean is a sub-optimal estimator of the true mean when the data distribution is heavy-tailed. For random $\rdv{X} \sim \ddist$ on $\RR$, assuming finite variance, setting $\theta^{\ast} = \exx_{\ddist}\rdv{X}$, we have
\begin{align*}
\exx_{\ddist}\left(\theta^{\ast}-\rdv{X}\right)^{2} \leq \exx_{\ddist}\left(\theta-\rdv{X}\right)^{2}, \quad \forall \, \theta \in \RR.
\end{align*}
That is, the mean minimizes the expected squared deviations from $\rdv{X}$. Given an independent random sample $\rdv{X}_{1},\ldots,\rdv{X}_{n}$, let $\widehat{\rdv{X}}_{n}$ denote an arbitrary estimator of $\exx_{\ddist}\rdv{X}$. Under just the finite variance assumption, writing $\sigma_{\ddist}^{2} \defeq \vaa_{\ddist}\rdv{X}$, ideally we would like to have ``sub-Gaussian'' guarantees of the form
\begin{align}\label{eqn:mean_est_subgaussian}
\prr\left\{ \abs{\widehat{\rdv{X}}_{n} - \exx_{\ddist}\rdv{X}} \leq a \sigma_{\ddist} \sqrt{\frac{\log(1/\delta)}{n}} \right\} \geq 1-\delta
\end{align}
for a sufficiently wide range of $\delta \in (0,1)$, where $a$ is a constant free of $\ddist$, $n$, and $\delta$; for more on this topic, see \citet{devroye2016a}. Unfortunately, it is well-known that under such a general setting, for small values of $a$ (e.g., $a=1$), the usual sample mean fails to satisfy such a property \citep[Thm.~1]{lugosi2019b}. The sample mean is of course an ERM solution under the squared error, namely
\begin{align*}
\overbar{\rdv{X}}_{n} \in \argmin_{\theta \in \RR} \frac{1}{n} \sum_{i=1}^{n} \left( \theta - \rdv{X}_{i} \right)^{2}, \qquad \overbar{\rdv{X}}_{n} \defeq \frac{1}{n} \sum_{i=1}^{n} \rdv{X}_{i}.
\end{align*}
If we consider these squared errors as losses $\ell(\theta;\rdv{X}_{i}) = (\theta - \rdv{X}_{i})^{2}$, then a well-established approach to obtain a more robust estimator is to modify the loss function to have sub-quadratic growth in the limit. A typical approach is to choose an estimator $\widehat{\rdv{X}}_{n}$ satisfying
\begin{align}\label{eqn:mean_est_Mest}
\widehat{\rdv{X}}_{n} \in \argmin_{\theta \in \RR} \frac{1}{n} \sum_{i=1}^{n} \rho\left(\frac{\theta - \rdv{X}_{i}}{s}\right)
\end{align}
where $\rho:\RR \to [0,\infty)$ is convex, approximately quadratic around zero, but with linear growth in the limit ($\rho(x)=\bigO(x)$ as $\abs{x} \to \infty$), and $s > 0$ is a scaling parameter used to control the bias incurred by using a non-quadratic loss function (larger $s$ means smaller bias). This basic ``M-estimation'' idea of course underlies many core techniques in robust statistics \citep{huber2009a}, and an important study from \citet{catoni2012a} showed how with appropriate settings of $s$, estimators given by (\ref{eqn:mean_est_Mest}) can be shown to satisfy (\ref{eqn:mean_est_subgaussian}) with nearly optimal constant $a$. Many other robust mean estimators satisfying the sub-Gaussian property (\ref{eqn:mean_est_subgaussian}) have since been studied, including the re-discovered mean-of-means \citep{lerasle2011a}, the well-known trimmed mean \citep{lugosi2019a}, and others; see \citet{lugosi2019b} for an up-to-date survey.

Since we know that there are practical algorithms for achieving the sub-Gaussian error bounds of the form (\ref{eqn:mean_est_subgaussian}), it is natural to consider such algorithms as \emph{sub-routines} that let us modify the naive ERM given by (\ref{eqn:classic_erm_condition}). That is, going back to our general setup described in \S{\ref{sec:intro}}, with $\rdv{Z} \sim \ddist$ and $\Z_{n}$ satisfying (\ref{eqn:iid_training_data}), writing $\loss(h) \defeq \ell(h;\rdv{Z})$ and $\loss_{i}(h) \defeq \ell(h;\rdv{Z}_{i})$ now for an arbitrary ``base'' loss function $\ell(\cdot)$, at the very least we know that pointwise in $h \in \HH$, we can construct an estimator using
\begin{align}\label{eqn:erm_Mest}
\widehat{\rdv{C}}_{n}(h) \in \argmin_{\theta \in \RR} \frac{1}{n} \sum_{i=1}^{n} \rho\left(\frac{\theta - \loss_{i}(h)}{s}\right)
\end{align}
which letting $\sigma_{\ddist}^{2}(h) \defeq \vaa_{\ddist}\loss(h)$, we know satisfies
\begin{align}\label{eqn:erm_subgaussian}
\prr\left\{ \abs{\widehat{\rdv{C}}_{n}(h) - \exx_{\ddist}\loss(h)} \leq a \sigma_{\ddist}(h) \sqrt{\frac{\log(1/\delta)}{n}} \right\} \geq 1-\delta.
\end{align}
This suggests the idea of replacing naive ERM algorithms (\ref{eqn:classic_erm_condition}) with alternatives of the form
\begin{align}\label{eqn:brownlees_condition}
\rdv{H}_{n} \in \argmin_{h \in \HH} \widehat{\rdv{C}}_{n}(h)
\end{align}
where $\widehat{\rdv{C}}_{n}(h)$ is computed using (\ref{eqn:erm_Mest}), or any of the other robust mean estimation procedures mentioned earlier. This novel approach to designing a learning algorithm was studied in seminal work by \citet{brownlees2015a}. This work is conceptually important because it rigorously demonstrates the statistical merits of using biased, robust estimators as objective functions. Statistically speaking, for learning tasks under heavy-tailed data, it is possible to show that any solution satisfying (\ref{eqn:brownlees_condition}) enjoys strong guarantees \emph{in the classical paradigm}, i.e., in terms of the expected loss condition (\ref{eqn:classic_pac_condition}), that are not possible for naive ERM (\ref{eqn:classic_erm_condition}).

Of course, we can also view the objective in (\ref{eqn:brownlees_condition}) from the population level as well, namely we can define a learning criterion class as follows:
\begin{align}
\crit_{\ddist}(h;s) \defeq \argmin_{\theta \in \RR} \exx_{\ddist}\rho\left(\frac{\theta - \loss(h)}{s}\right).
\end{align}
With proper choice of $\rho$ and $s > 0$, one can control the bias such that good performance in terms of $\crit_{\ddist}(h;s)$ implies good performance \emph{on average}. More generally, however, it is perfectly plausible to fix $s$ and use $\crit_{\ddist}(h;s)$ as the primary criterion for performance. In any case, the chief limitation of this approach is computational; actual computation of $\rdv{H}_{n}$ in (\ref{eqn:brownlees_condition}) is left abstract, since the minimization of $\widehat{\rdv{C}}_{n}(\cdot)$ given by (\ref{eqn:erm_Mest}) is a challenging optimization problem.\footnote{Since the M-estimator is itself defined as the solution of a minimization problem, minimizing that quantity as a function of $h \in \HH$ is what is termed a \term{bi-level program}.} Since M-parameters (including all quantiles) are easy to compute but difficult to minimize directly, they are perhaps best suited as \emph{evaluation} metrics to be coupled with a more convenient surrogate objective \citep{holland2017a}. For direct minimization, a natural alternative is that of criteria which offer a smooth approximation to quantiles, which are easy to optimize under typical base loss functions. Exponential smoothing or ``tilting'' of the loss distribution \citep{li2020a,li2021a} can be used effectively to this end, and ``inverted'' variants of CVaR \citep{lee2020a} can also be used to obtain a similar effect. We will revisit both of these examples in more detail in \S{\ref{sec:lcrit_classes}}.

\subsection{Variance Control}\label{sec:core_concepts_variance}

All else equal, smaller variance tends to make statistical inference easier. Once again we will use one-dimensional mean estimation as a motivating example, retaining the notation from \S{\ref{sec:core_concepts_robustness}}, as well as the assumption (\ref{eqn:iid_training_data}). Assuming $\rdv{X} \sim \ddist$ has finite variance $\sigma_{\ddist}^{2}$, Chebyshev's inequality gives us a simple error bound on the sample mean $\overbar{\rdv{X}}_{n}$, namely that
\begin{align}\label{eqn:mean_est_chebyshev}
\abs{\overbar{\rdv{X}}_{n}-\exx_{\ddist}\rdv{X}} \leq \sigma_{\ddist}\sqrt{\frac{1}{n\delta}}
\end{align}
holds with probability no less than $1-\delta$, for any choice of $0 < \delta \leq 1$. If we further assume that $\rdv{X}$ takes values on the unit interval $[0,1] \subset \RR$ with probability one, then once again with probability no less than $1-\delta$, we have
\begin{align}\label{eqn:mean_est_bennett}
\abs{\overbar{\rdv{X}}_{n}-\exx_{\ddist}\rdv{X}} \leq \sigma_{\ddist}\sqrt{\frac{2\log(2/\delta)}{n}} + \frac{\log(2/\delta)}{3n}.
\end{align}
This follows from a result known as Bennett's inequality.\footnote{See \citet[Ch.~2]{boucheron2013a} for an authoritative modern reference.} Both (\ref{eqn:mean_est_chebyshev}) and (\ref{eqn:mean_est_bennett}) show us that all else equal, a smaller variance can enable sharper guarantees. With this in mind, recalling the ERM condition (\ref{eqn:classic_erm_condition}) and the fact that in general we may have many ERM solutions, it is natural to prefer those with smaller variance (in terms of the loss distribution). Indeed, as the bound (\ref{eqn:mean_est_bennett}) suggests, it is well-established that achieving a small variance can be used to guarantee ``fast rates'' (i.e., better than $\bigO(\sqrt{1/n})$) for empirical risk minimizers.\footnote{See for example \citet{duchi2019a} and the references therein.} In terms of designing learning criteria, the simplest choice is just the sum
\begin{align}\label{eqn:mean_stdev}
\crit_{\ddist}(h) = \exx_{\ddist}\loss(h) + \sqrt{\sigma_{\ddist}^{2}(h)}.
\end{align}
This property of the loss distribution dates back to classical work on portfolio optimization by \citet{markowitz1952a} (originally without the square root on $\sigma_{\ddist}^{2}(h)$), and replacing $\ddist$ with the empirical distribution induced by $\Z_{n}$, solutions to the resulting learning criterion minimization were analyzed in influential work by \citet{maurer2009a}. While statistically easy to interpret and advantageous in theory, much like the robust objective (\ref{eqn:erm_Mest}) from \S{\ref{sec:core_concepts_robustness}}, naively trying to optimize for (\ref{eqn:mean_stdev}) can be computationally challenging. In particular, even when the loss $h \mapsto \ell(h;\rdv{Z})$ is convex, we cannot in general guarantee that $h \mapsto \crit_{\ddist}(h)$ in (\ref{eqn:mean_stdev}) is convex. Recently, important insights have been established linking mean-variance to certain objective functions from the field of \term{distributionally robust optimization (DRO)} \citep{duchi2019a,gotoh2018a}, in which the goal is to minimize the worst-case expected loss to be incurred over a \emph{set} of distributions, rather than just one pre-specified $\ddist$. These methods use asymmetric measures of deviation, which place more weight on worst-case tails and preserve convexity. This DRO approach to stochastic optimization also has important applications to deal with data distribution shift, and will be introduced formally in \S{\ref{sec:lcrit_classes}}.

\subsection{Risk-Averse Learning}\label{sec:core_concepts_riskaverse}

In contrast to robustness concerns discussed in \S{\ref{sec:core_concepts_robustness}}, where the expected loss was too sensitive to the loss tails to allow for high-confidence guarantees, sometimes the tails are precisely what we want to prioritize. This is an example of \term{risk aversion}, a tendency to overweight bad events and underweight good events, where the terms ``bad'' and ``good'' may be subjective or have different meanings depending on the task at hand.\footnote{A classical reference from economics is \citet{pratt1964a}. See also \citet{tversky1992a}.} One particularly intuitive way to define a ``bad'' event is to use the quantiles of the loss distribution as a threshold. Formally, let us denote the distribution function of $\loss$ by $\dfun_{\ddist}(u) \defeq \prr\left\{ \loss \leq u\right\}$, and using this, we define the $\beta$-level \term{quantile} of $\loss$ as follows:\footnote{When $\dfun_{\ddist}(\cdot)$ is not continuous, $\quant_{\beta}$ in (\ref{eqn:quantile_population}) is sometimes distinguished as the ``left'' or ``lower'' quantile.}
\begin{align}\label{eqn:quantile_population}
\quant_{\beta}(\loss) \defeq \min \left\{ u \in \RR: \dfun_{\ddist}(u) \geq \beta \right\}, \qquad 0 < \beta < 1.
\end{align}
Since losses come with the interpretation of larger being worse, a natural ``bad'' event is something like $\{\loss \geq \quant_{0.95}(\loss)\}$. As such, the average loss incurred beyond this threshold is a perfectly natural class of learning criteria that generalizes beyond the mean:
\begin{align}\label{eqn:cvar_thresholded}
\crit_{\ddist}(h;\beta) = \exx_{\ddist}\left[ \loss(h)\indic\{ \loss(h) \geq \quant_{\beta}(h) \} \right], \quad 0 \leq \beta < 1.
\end{align}
Once again, in (\ref{eqn:cvar_thresholded}) we overload our formal notation $\quant_{\beta}$ by defining $\quant_{\beta}(h) \defeq \quant_{\beta}(\loss(h))$, with the special case of $\beta = 0$ defined as
\begin{align*}
\crit_{\ddist}(h;0) \defeq \lim\limits_{\beta \to 0_{+}} \crit_{\ddist}(h;0) = \exx_{\ddist}\loss(h).
\end{align*}
Given a training sample $\Z_{n} = (\rdv{Z}_{1},\ldots,\rdv{Z}_{n})$ and sorting the training losses as before such that $\loss_{[1]} \geq \ldots \geq \loss_{[n]}$, for any quantile level $0 \leq \beta \leq 1$, there exists a $k \in [n]$ such that the empirical objective analogous to (\ref{eqn:cvar_thresholded}) can be written
\begin{align*}
\widehat{\rdv{C}}_{n}(h;k) = \frac{1}{n} \sum_{i=1}^{k}\loss_{[i]}(h).
\end{align*}
A re-scaled variant known as the ``average top-$k$'' objective is given as
\begin{align}\label{eqn:average_top_k_empirical_risk}
\frac{n}{k}\widehat{\rdv{C}}_{n}(h;k) = \frac{1}{k} \sum_{i=1}^{k} \loss_{[i]}(h)
\end{align}
and algorithms based on (\ref{eqn:average_top_k_empirical_risk}) have been studied by \citet{shalev2016a} as well as \citet{fan2017b,fan2017a}, though the issues of robustness mentioned in \S{\ref{sec:core_concepts_robustness}} are of course still relevant here, which has naturally led to robustified variants \citep{holland2021c}. Setting $k=n$ returns the usual empirical mean, and $k=1$ yields a maximum value. Assuming the distribution function $\dfun_{\ddist}$ is continuous, we have $\prr\{\loss(h) > \quant_{\beta}(h) \} = \prr\{\loss(h) \geq \quant_{\beta}(h) \} = 1-\beta$, and we can easily construct a (population) criterion corresponding to (\ref{eqn:average_top_k_empirical_risk}) as
\begin{align}\label{eqn:cvar_population}
\crit_{\ddist}^{\text{CVaR}}(h;\beta) \defeq \frac{\crit_{\ddist}(h;\beta)}{\prr\{\loss(h) \geq \quant_{\beta}(h) \}} = \frac{\exx_{\ddist}\loss(h)\indic\{ \loss(h) \geq \quant_{\beta}(h) \}}{1-\beta} = \exx_{\ddist}\left[ \loss(h) \cond \loss(h) \geq \quant_{\beta}(h) \right].
\end{align}
In both (\ref{eqn:average_top_k_empirical_risk}) and (\ref{eqn:cvar_population}), we are clearly considering the mean of a pre-specified fraction of the largest losses. This quantity has been called the \term{expected shortfall} \citep{acerbi2002b}, and it is also well-known as \term{conditional value-at-risk (CVaR)} \citep{rockafellar2000a}, a natural name given the right-most form in (\ref{eqn:cvar_population}).\footnote{Nomenclature in the financial risk literature is somewhat convoluted; the paper of \citet{acerbi2002c} is a useful reference. For extensions beyond the continuous case, see \citet{rockafellar2002a}.} The quantity given in (\ref{eqn:cvar_population}) can be seen as a special case of a much larger class of criteria under the name \term{optimized certainty equivalent (OCE)}, and is in fact also a limiting case of DRO risk functions \citep{duchi2018b}. Both will be introduced with more generality in \S{\ref{sec:lcrit_classes}}.

\subsection{Fairness-Aware Learning}\label{sec:core_concepts_fairness}

Since machine learning methods are increasingly being utilized to automate or support real-world decisions, it is natural to run into issues of ethics and in particular \emph{fairness} to individuals or groups with certain traits. To give this broad concept an intuitive and clear formalization, assuming the underlying data $\rdv{Z} \sim \ddist$ is a tuple of one or more random variables, let $\rdv{S}$ denote a ``sensitive'' variable, representing one or more of the random variables composing $\rdv{Z}$. In the context of supervised learning, where $\rdv{Z} = (\rdv{X},\rdv{Y})$ and we want to choose some $h \in \HH$ such that we reliably get a sufficiently good prediction $h(\rdv{X}) \approx \rdv{Y}$, the sensitive variable $\rdv{S}$ might be a subset of features in $\rdv{X}$, or indeed the class label $\rdv{Y}$. For example, say $\rdv{Y}$ is binary-valued and indicates success at graduating from university, and $\rdv{X}$ represents various attributes of an individual such as test grades in primary and secondary school, extra-curricular honors, gender, ethnicity, and family income. In this case, $\rdv{S}$ might correspond to gender, ethnicity, and family income. If we take the traditional approach of optimizing for prediction accuracy (i.e., minimizing $\prr\{h(\rdv{X}) \neq \rdv{Y}\}$), then it is of course plausible that $\rdv{S}$ plays a key role in accurately predicting $\rdv{Y}$, but without intervention, this could result in individuals with certain traits being assigned predictions that are intuitively unfair, even if they are ``optimal'' in terms of accuracy.

To mitigate this tendency, we must either change the ultimate objective directly, or enforce some fairness constraints. There are numerous specific examples of fairness constraints in the literature, but one of the strongest is (statistical) \term{parity} \citep{dwork2012a}, which asks that $h(\rdv{X})$ and $\rdv{S}$ be statistically independent. This is the same as requiring that $h(\rdv{X})$ and $h(\rdv{X}) \cond \rdv{S}$ have the same distribution. A closely related notion looks at \emph{performance} through the lens of a loss function, i.e., asking that $\ell(h;\rdv{Z})$ and $\ell(h;\rdv{Z}) \cond \rdv{S}$ have the same distribution. Note that this lets us go well beyond the supervised learning paradigm.\footnote{Consider the example of \citet{samadi2018a}, where dimension reduction using standard PCA can be unfair in that low-education sub-groups incur a disproportionately high reconstruction error.} Asking for the marginal and conditional distributions to be identical is an extremely strong requirement; a much weaker constraint just asks that their expected values align, i.e., that
\begin{align}\label{eqn:fairness_subgroup_risk}
\exx_{\ddist}\ell(h;\rdv{Z}) = \exx_{\ddist}\left[ \ell(h;\rdv{Z}) \cond \rdv{S} \right]
\end{align}
hold with probability 1 over the random draw of $\rdv{S}$. This notion of ``fairness via sub-group risks'' is formulated in a lucid way by \citet{williamson2019a}, though as they mention, the underlying notion is present in several works preceding theirs (e.g., \citet{hashimoto2018a}). While (\ref{eqn:fairness_subgroup_risk}) is weaker than asking the marginal and conditional distributions to align, the goal of choosing $h \in \HH$ such that this equality holds is still not very realistic, and further relaxations are required to link fairness constraints up to practical learning algorithms.

In this direction, note that the right-hand side of (\ref{eqn:fairness_subgroup_risk}) is a random variable depending on the draw of $\rdv{S}$, but the left-hand side is a constant. As such, for (\ref{eqn:fairness_subgroup_risk}) to hold, the random variable on the right-hand side must have zero variance: $\vaa_{\rdv{S}}(\exx_{\ddist}\left[ \ell(h;\rdv{Z}) \cond \rdv{S} \right]) = 0$. Relaxing this, we could ask for $\varepsilon$-small variance, i.e., that $\vaa_{\rdv{S}}(\exx_{\ddist}\left[ \ell(h;\rdv{Z}) \cond \rdv{S} \right]) \leq \varepsilon$ holds for some $\varepsilon > 0$ that we specify. Recalling our previous discussion of the benefits of small variance in \S{\ref{sec:core_concepts_variance}}, as a direct approach to this problem, we could naively introduce a learning criterion of the form
\begin{align}
\crit_{\ddist}(h) = \exx_{\ddist}\loss(h) + \sqrt{\vaa_{\rdv{S}}\left( \exx_{\ddist}\left[ \loss(h) \cond \rdv{S} \right] \right)}
\end{align}
where once again we write $\loss(h) \defeq \ell(h;\rdv{Z})$ for readability. Just as in \S{\ref{sec:core_concepts_variance}}, using the symmetric variance as-is leads to computational difficulties, and more congenial alternatives such as DRO and CVaR have been put to effective use in this direction \citep{hashimoto2018a,williamson2019a}. A full description of these classes will be given in \S{\ref{sec:lcrit_classes}} to follow.

\section{Key Classes of Learning Criteria}\label{sec:lcrit_classes}

With the general learning problem formulated in \S{\ref{sec:intro_our_view}} coupled with the motivating examples discussed in \S{\ref{sec:core_concepts}}, we will now dive into an exposition of some of the most important classes of learning criteria that have arisen in and around the machine learning literature. For readability, we shall divide this section into several sub-sections, each of which dedicated to a certain class of interest.

\subsection{Order-Based Criteria}\label{sec:lcrit_classes_orderbased}

Arguably the most intuitive strategy to design criteria going beyond the usual expected value is to sort the (empirical) losses and assign weights depending on rank before summation. Sorting can be done in either ascending or descending order, as long as we are consistent with the weighting. We already saw examples of losses sorted in descending order in \S{\ref{sec:intro_related_work}} and \S{\ref{sec:core_concepts_riskaverse}}, and here we will consider sorting in ascending order, since that aligns us with the literature we take as context. With $n$ observations, let $\loss_{(i)}(h)$ denote the $i$th-smallest loss under $h$. That is, the losses are sorted such that $\loss_{(1)}(h) \leq \ldots \leq \loss_{(n)}(h)$ holds. Letting $\mathbf{w} \defeq (w_{1},\ldots,w_{n})$ denote non-negative weights, we can define a new objective function as follows:
\begin{align}\label{eqn:quantile_risk_empirical}
\widehat{\rdv{C}}_{n}(h;\mathbf{w}) = \sum_{i=1}^{n}w_{i}\loss_{(i)}(h).
\end{align}
One classical motivation for this is to discard errant values to aid in estimating the population location, typically the mean or median.\footnote{This is the ``discard-average'' in classic work by \citet{daniell1920a}, who also cites a 1912 work of Poincar\'{e}.} This can be achieved in a strict sense by setting $w_{i} = 0$ for rank $i$ above or below certain thresholds, such that a pre-specified fraction of the data is discarded. More generally, the idea of setting weights as a function of rank is an important idea with a long history in statistics.\footnote{See for example \citet{bickel1975} and the references therein. A more modern take on the optimality of trimmed-means is given by \citet{lugosi2019a}.} Making this a bit more explicit, letting $W:[0,1] \to [0,\infty)$ denote a weighting function, the weights in (\ref{eqn:quantile_risk_empirical}) can be set as $w_{i} = W(i/n)$, for each $i \in [n]$. Denoting the empirical distribution function of the losses by $\rdv{F}_{n}(u;h) \defeq \sum_{i=1}^{n}\indic\{\loss_{i}(h) \leq u\}/n$, note that we have $\rdv{F}_{n}(\loss_{(i)}(h);h) = i/n$, and thus (\ref{eqn:quantile_risk_empirical}) can be re-written as follows:
\begin{align}\label{eqn:quantile_risk_empirical_cdfform}
\widehat{\rdv{C}}_{n}(h;\mathbf{w}) = \sum_{i=1}^{n}W(\rdv{F}_{n}(\loss_{(i)}))\loss_{(i)} = \sum_{i=1}^{n}W(\rdv{F}_{n}(\loss_{i}))\loss_{i}
\end{align}
where we have suppressed the dependence on $h$ in the notation just for readability. Reading left to right, the second equality in (\ref{eqn:quantile_risk_empirical_cdfform}) clearly holds since passing unsorted losses through the empirical CDF ensures they get the correct rank-based weighting. In moving from the preceding empirical objective to (population) learning criteria, the two forms given in (\ref{eqn:quantile_risk_empirical_cdfform}) are indicative of two paths we can take. Since both paths are meaningful and have relevant literature, we will look at both of them, one at a time.

The first path is centered around sorted losses. To link up sorted losses to quantiles, recall our (population) definition of $\quant_{\beta}(\loss)$ in (\ref{eqn:quantile_population}), and replace the true distribution function $\dfun_{\ddist}$ with its empirical analogue $\rdv{F}_{n}$. This yields the empirical quantile (viewed as a function of $\beta$) as follows:
\begin{align}\label{eqn:quantile_empirical}
\rdv{Q}_{\beta,n} \defeq \min\left\{ u \in \RR: \rdv{F}_{n}(u) \geq \beta \right\}, \quad 0 < \beta \leq 1.
\end{align}
Ranked losses have a clear relation to the empirical quantiles. Note that for any $0 < \beta \leq 1$, we have $\rdv{Q}_{\beta,n} = \loss_{(\lceil n\beta \rceil)}$. Since $\beta \mapsto \rdv{Q}_{\beta,n}(\loss)$ is just a simple step function, if we assume that $W(\beta) \geq 0$ for $\beta \in \{1/n,2/n,\ldots,1\}$ and $W(\beta) = 0$ elsewhere, we can once again re-write (\ref{eqn:quantile_risk_empirical}) in terms of quantiles using an integral form, namely
\begin{align}\label{eqn:quantile_risk_empirical_intform}
\widehat{\rdv{C}}_{n}(h;\mathbf{w}) = \int_{0}^{1} W(\beta)\rdv{Q}_{\beta,n}(\loss) \, \dif{\beta}
\end{align}
where we set $w_{i} = W(i/n)$ for each $i \in [n]$. The equality in (\ref{eqn:quantile_risk_empirical_intform}) uses the fact that $\rdv{F}_{n}(\rdv{Q}_{\beta,n}(\loss))=\beta$ for all $\beta \in \{1/n,2/n,\ldots,1\}$, plus the first equality in (\ref{eqn:quantile_risk_empirical_cdfform}). With this (empirical) form in mind, a population variant follows:
\begin{align}\label{eqn:quantile_risk_pop_intform}
\crit_{\ddist}(\loss;W) \defeq \int_{0}^{1} W(\beta)\quant_{\beta}(\loss) \, \dif{\beta}.
\end{align}
Clearly, the criterion in (\ref{eqn:quantile_risk_pop_intform}) is completely characterized by the weighting function $W(\cdot)$. It is also natural to extend the criterion definition to allow for any function $W(\cdot)$ such that the integral on the right-hand side of (\ref{eqn:quantile_risk_pop_intform}) is finite (or at least well-defined). The general form given by (\ref{eqn:quantile_risk_pop_intform}) can be used to capture a wide variety of well-known properties. Linear combinations of order statistics, also known as ``L-estimates,'' have played an important role in robust statistics \citep{huber2009a}. Applications to robust unsupervised learning, replacing average distortion with an ``L-statistic'' have also been studied recently by \citet{maurer2020a}.

The second path, motivated by the right-most form in (\ref{eqn:quantile_risk_empirical_cdfform}), places its emphasis on rank-based weighting (via a $W$-transformed distribution function) applied to \emph{unsorted} losses. Learning criteria of the form
\begin{align}\label{eqn:quantile_risk_pop_expform}
\crit_{\ddist}(\loss;W) = \exx_{\ddist}W(\dfun_{\ddist}(\loss))\loss
\end{align}
play the central role here. This form has intuitive appeal, since we are working with losses rather than quantiles, and concentration properties of the empirical CDF are well-known. That said, we cannot naively cast this as a traditional ERM-style learning problem. This is because for any $i \in [n]$, the empirical quantity $W(\rdv{F}_{n}(\loss_{i}))\loss_{i}$ depends on the entire sample $\Z_{n}$, and thus independence is lost. Getting around this requires $W(\cdot)$ to satisfy certain continuity properties, and sharp concentration of $\rdv{F}_{n}(\cdot;h)$ around $\dfun_{\ddist}(\cdot;h)$, uniform over $h \in \HH$. Conditions for the uniform convergence of (\ref{eqn:quantile_risk_empirical_cdfform}) to $\crit_{\ddist}(h;W)$ in (\ref{eqn:quantile_risk_pop_expform}) were studied by \citet{khim2020a} under bounded losses and Lipschitz continuous $W(\cdot)$, with $\crit_{\ddist}(\cdot)$ called the ``L-risk.'' An extension centered around H{\" o}lder-continuous functions defined on families of distribution functions was developed recently by \citet{liu2022a}. Finally, it should be noted that when the underlying distribution function $\dfun_{\ddist}$ is invertible, the two paths coincide, i.e., $\crit_{\ddist}(\loss;W)$ given in (\ref{eqn:quantile_risk_pop_intform}) and (\ref{eqn:quantile_risk_pop_expform}) are equal \citep[Lem.~1]{holland2022b}.

To complement the preceding general exposition, we look at several prominent special cases that arise in the machine learning literature.
\begin{ex}[CVaR]\label{ex:cvar_avar_form}
It is well-established that CVaR as seen in (\ref{eqn:cvar_population}) is a special case of the form (\ref{eqn:quantile_risk_pop_intform}). In particular, we have
\begin{align}\label{eqn:cvar_avar_form}
\crit_{\ddist}^{\text{CVaR}}(h;\beta) = \int_{0}^{1} W_{\beta}(u) \quant_{u}(h) \, \dif{u}, \quad W_{\beta}(u) \defeq \frac{\indic\{u \geq \beta\}}{1-\beta}
\end{align}
for any choice of $0 \leq \beta < 1$.\footnote{See for example Corollary 4.3 and Eqn.~3.3 of \citet{acerbi2002c}, or \citet[Eqn.~6]{shapiro2013a}.} Since this weighting function determines a probability measure over the unit interval, we are ``averaging the quantile,'' and since the left quantile is often referred to as the \term{value-at-risk}, CVaR is also sometimes referred to as the \term{average value-at-risk (AVaR)} \citep{ruszczynski2006a}. In the context of machine learning, CVaR is perhaps the best-known alternative to the mean. An early application can be found in \citet{kashima2007a}, but analysis of stochastic learning algorithms under CVaR is far more recent; see \citet{cardoso2019a,soma2020a,curi2020a,holland2021c} for representative works.\hfill$\blacksquare$
\end{ex}

\begin{ex}[Spectral risk]\label{ex:spectral_risk}
A natural extension of the form (\ref{eqn:cvar_avar_form}) is to consider a weighting function that places progressively more weight on the upper tails, satisfying
\begin{align}\label{eqn:spectral_weight_fn}
W(\cdot) \text{ is non-decreasing on } [0,1] \text{ and } \int_{0}^{1} W(u) \, \dif{u} = 1.
\end{align}
Under the specifications of (\ref{eqn:spectral_weight_fn}), learning criteria of the form (\ref{eqn:quantile_risk_pop_intform}) are called \term{spectral risks} \citep{acerbi2002a}. Spectral risk estimators are studied by \citet{pandey2019a}, and learning theory for algorithms based on spectral risks is developed by \citet{holland2022b} and \citet{liu2022a}.\hfill$\blacksquare$
\end{ex}

\begin{ex}[Average top-$k$]\label{ex:average_topk}
Given a finite sample $\Z_{n} = (\rdv{Z}_{1},\ldots,\rdv{Z}_{n})$ and utilizing a weight function defined as
\begin{align*}
W_{k}(u) \defeq \frac{\indic\{u \geq (n-k)/n\}}{(k/n)}
\end{align*}
ensures we cut off all losses below a certain rank $k \in [n]$, and then re-scale to get the arithmetic mean of the $k$ largest losses. Spelling this all out a bit more explicitly, setting the weight function to $W=W_{k}$ as just described in (\ref{eqn:quantile_risk_empirical_intform}), we obtain
\begin{align*}
\int_{0}^{1} W_{k}(u)\rdv{Q}_{u,n}(h) \, \dif{u} = \left(\frac{n}{k}\right)\int_{(n-k)/n}^{1}\rdv{Q}_{u,n}(h) \, \dif{u} = \left(\frac{n}{k}\right)\sum_{i=n-k+1}^{n}\frac{\loss_{(i)}(h)}{n} = \frac{1}{k} \sum_{i=1}^{k}\loss_{[i]}(h).
\end{align*}
That is, we recover the average top-$k$ empirical risk (\ref{eqn:average_top_k_empirical_risk}) as seen earlier in \S{\ref{sec:core_concepts_riskaverse}}. As mentioned before, algorithms optimizing this objective has been studied in the recent machine learning literature \citep{fan2017b,fan2017a,shalev2016a}. A handy computational fact is that the map $(x_1,\ldots,x_n) \mapsto \sum_{i=1}^{k}x_{[i]}$ is convex \citep{ogryczak2003a}, although in general it is not differentiable. For smooth approximations, see Example \ref{ex:tilted_risk} in the next sub-section.\hfill$\blacksquare$
\end{ex}

\subsection{OCE Risks}\label{sec:lcrit_classes_oce}

We proceed by looking at a class of learning criteria with motivations quite different from the order-based criteria seen in the preceding \S{\ref{sec:lcrit_classes_orderbased}}. To provide this class with some appropriate historical context, let $\rdv{X} \sim \ddist$ be a random variable representing some uncertain future ``payout'' (larger is better), and let $f:\RR \to \RR$ be in charge of assigning ``utility'' values to raw payouts. Decision making under uncertainty is traditionally centered around the expected utility, namely $\exx_{\ddist}f(\rdv{X})$. This fact is of course closely related to the classical paradigm of statistical learning theory discussed in \S{\ref{sec:intro_classical}}.\footnote{For more on expected utility and decision-making under certainty, see \S{\ref{sec:history_expectation}}.} With these elements in place, the \term{certainty equivalent} of $\rdv{X}$ under $f$ is formally defined as
\begin{align}\label{eqn:cert_equivalent}
\text{CE}_{\ddist}(\rdv{X};f) \defeq \left\{ \theta \in \RR: f(\theta) = \exx_{\ddist}f(\rdv{X}) \right\}.
\end{align}
In words, from the perspective of $f(\cdot)$, receiving $\theta \in \text{CE}_{\ddist}(\rdv{X};f)$ with certainty is just as appealing as receiving $\rdv{X}$ with uncertainty, where the uncertainty is summarized by averaging.\footnote{See \citet[Defn.~6.C.2]{mascolell1995MicroEcon} for more background on this notion in the context of preferences.} Building upon this concept, if $\theta \in \RR$ represents a certain payment, there are many situations in which committing to $\theta$ leaves us with an uncertain payment of $\rdv{X}-\theta$. When $\theta$ is something we can control, it is natural to ``optimize'' this value such that the overall outcome is as preferable as possible, at least on average. Arguably the simplest way to evaluate such a situation is purely in terms of expected utility, with the best possible outcome quantified as
\begin{align*}
\sup_{\theta \in \RR} \left[ f(\theta) + \exx_{\ddist}f(\rdv{X}-\theta)\right].
\end{align*}
A slightly modified version has us evaluate certain outcomes in their raw units, and uncertain outcomes using utility, taking the expectation of their sum. More explicitly, we define
\begin{align}\label{eqn:oce_utility}
\overbar{\text{OCE}}_{\ddist}(\rdv{X};f) \defeq \sup_{\theta \in \RR} \left[ \theta + \exx_{\ddist}f(\rdv{X}-\theta) \right]
\end{align}
known as the \term{optimized certainty equivalent (OCE)}, originally proposed and studied by \citet{bental1986a} assuming that $f(\cdot)$ is concave, monotonic (non-decreasing), and ``normalized'' such that $f(0) = 0$ and $1 \in \partial{f}(0)$.\footnote{See \citet[Defn.~2.1]{bental2007a} for example.} Regarding the naming, (\ref{eqn:oce_utility}) was originally called NCE (new certainty equivalent) in \citep{bental1986a}, but two decades later, the authors published an updated work on this quantity \citep{bental2007a}, in which it was re-named OCE.\footnote{Still, the naming is peculiar; one would expect the $\argmax_{\theta \in \RR}$ rather than the $\sup_{\theta \in \RR}$ to be called OCE.}

Shifting our mindset from positively-oriented ``payouts'' to negatively oriented losses gives us an analogous property for loss distributions. Letting $\loss$ represent a random loss as usual, the \term{OCE risk} associated with $\loss$ under $\phi$ is defined by
\begin{align}\label{eqn:oce_risk}
\underline{\text{OCE}}_{\ddist}(\loss;\phi) \defeq \inf_{\theta \in \RR} \left[ \theta + \exx_{\ddist}\phi(\loss-\theta) \right]
\end{align}
where $\phi:\RR \to \RR$ is assumed to have the same properties as $f$ in (\ref{eqn:oce_utility}), except that $\phi$ is convex rather than concave, but still non-decreasing. We can easily link the OCE risk of (\ref{eqn:oce_risk}) with the OCE for payouts given by (\ref{eqn:oce_utility}); given a utility function $f$, a ``disutility'' function is naturally created by taking $\phi$ such that $\phi(x) = -f(-x)$. Note that there are two minus signs here. First, $x$ passed to $\phi$ will be a loss, and thus since we need to pass $f$ a payout, we pass it $-x$. Second, since $f(\cdot)$ returns utility (larger is better), $-f(\cdot)$ is something to be minimized. This setting yields
\begin{align*}
\underline{\text{OCE}}_{\ddist}(\loss;\phi) & = \inf_{\theta \in \RR}\left[ \theta - \exx_{\ddist}f(-\loss+\theta) \right]\\
& = -\sup_{\theta \in \RR}(-1)\left[ \theta - \exx_{\ddist}f(-\loss+\theta) \right]\\
& = -\sup_{\theta \in \RR}\left[ -\theta + \exx_{\ddist}f(-\loss+\theta) \right]\\
& = -\sup_{\theta \in \RR}\left[ \theta + \exx_{\ddist}f(-\loss-\theta) \right]\\
& = -\overbar{\text{OCE}}_{\ddist}(-\loss;f)
\end{align*}
recalling that $\underline{\text{OCE}}_{\ddist}(\cdot)$ on the left-most side follows the definition (\ref{eqn:oce_risk}), and $\overbar{\text{OCE}}_{\ddist}(\cdot)$ on the right-most side follows (\ref{eqn:oce_utility}). Regardless of which form we use, the notion of a \emph{trade-off} between certain and uncertain outcomes is quite fundamental in the theory of risk functions.

If we overload the notation in (\ref{eqn:oce_risk}) setting $\underline{\text{OCE}}_{\ddist}(h;\phi) \defeq \underline{\text{OCE}}_{\ddist}(\loss(h);\phi)$ for each $h \in \HH$, this becomes another natural criterion for machine learning problems. One very important quality of this class is the ease of preserving \emph{convexity}. If $h \mapsto \loss(h) = \ell(h;\rdv{Z})$ is convex on $\HH$ with probability one, then $h \mapsto \underline{\text{OCE}}_{\ddist}(h;\phi)$ is also convex on $\HH$. Preserving convexity depends critically on the monotonicity of $\phi$; the convexity of $\phi$ alone is not sufficient.\footnote{See the ``Convexity Theorem'' of \citet{rockafellar2013a}.} The OCE risk class as a whole was tackled using traditional statistical learning theory techniques and assumptions in an elegant way by \citet{lee2020a}. Following their notation, by setting $\varphi(x) \defeq \phi(x)-x$, we can re-write any OCE risk as
\begin{align}\label{eqn:oce_risk_varphi}
\underline{\text{OCE}}_{\ddist}(\loss;\phi) = \exx_{\ddist}\loss + \inf_{\theta \in \RR}\left[ \exx_{\ddist}\varphi(\loss-\theta) \right]
\end{align}
which gives this learning criterion a natural interpretation as a ``mean + deviation'' sum.\footnote{Using terminology from the fundamental risk quadrangle of \citet{rockafellar2013a}, $\exx_{\ddist}\phi$ in (\ref{eqn:oce_risk}) is a ``regret measure,'' and $\exx_{\ddist}\varphi$ in (\ref{eqn:oce_risk_varphi}) is an ``error measure'' quantifying the ``nonzeroness'' of $\loss$.} With the assumptions placed on $\phi$ (and thus $\varphi$), it follows that $\varphi(x) \geq 0$ for all $x \in \RR$, and thus (\ref{eqn:oce_risk_varphi}) shows us that $\exx_{\ddist}\loss \leq \underline{\text{OCE}}_{\ddist}(\loss;\phi)$ always holds. As we will see in the examples to follow, OCE risks include important sub-classes that are well-known and well-studied in their own right.

\begin{ex}[Tilted risk]\label{ex:tilted_risk}
For $\gamma \neq 0$, let us define
\begin{align}\label{eqn:tilted_risk}
\crit_{\ddist}(h;\gamma) \defeq \frac{1}{\gamma}\log\left(\exx_{\ddist}\mathrm{e}^{\gamma\loss(h)}\right)
\end{align}
and call this the \term{$\gamma$-tilted risk}, a term used recently in work by \citet{li2020a,li2021b}. First and foremost, when $\gamma > 0$, this is a special case of OCE risk (\ref{eqn:oce_risk}), with $\phi_{\gamma}(x) \defeq (\mathrm{e}^{\gamma{x}}-1)/\gamma$. It also has a rather special property in that it can be seen as a certainty equivalent under $\phi_{\gamma}$, which is to say that for any choice of $\gamma \neq 0$, one can easily verify that
\begin{align}
\phi(\crit_{\ddist}(h;\gamma)) = \exx_{\ddist}\phi_{\gamma}(\loss(h)).
\end{align}
It must be noted however that when $\gamma < 0$, this $\crit_{\ddist}(\cdot;\gamma)$ is not a valid OCE risk, since $\phi_{\gamma}(\cdot)$ is concave, and plugging $\phi = \phi_{\gamma}$ into (\ref{eqn:oce_risk}) clearly leads to $\underline{\text{OCE}}_{\ddist}(\loss;\phi_{\gamma}) = -\infty$. That said, while the OCE form is meaningless when $\gamma < 0$, the tilted risk form (\ref{eqn:tilted_risk}) is perfectly valid, and for general choices of $\gamma \neq 0$, this objective has a long history of use in risk-sensitive reinforcement learning; see \citet[\S{3.2.1}]{garcia2015a}. As described by \citet{follmer2011a}, the special case of $\gamma > 0$ is often referred to as the \term{entropic risk} in financial contexts. Viewed as a function of $\gamma$, the quantity $\gamma\crit_{\ddist}(h;\gamma)$ is the logarithm of the \term{moment generating function} of $\loss(h)$, a quantity that plays an important role in statistical theory \citep[Ch.~3]{stuart1994KendallVol1}. Replacing $\ddist$ in (\ref{eqn:tilted_risk}) with the empirical distribution on $\Z_{n} = (\rdv{Z}_{1},\ldots,\rdv{Z}_{n})$, we get
\begin{align}\label{eqn:tilted_risk_empirical}
\widehat{\rdv{C}}_{n}(h;\gamma) \defeq \frac{1}{\gamma}\log\left(\frac{1}{n} \sum_{i=1}^{n}\mathrm{e}^{\gamma\loss_{i}(h)}\right)
\end{align}
and taking $\gamma > 0$ sufficiently large gives us a smooth approximation to the ``largest loss'' objective $\loss_{(n)} = \loss_{[1]}$, a well-known technique from the optimization literature \citep{pee2011a}.\hfill$\blacksquare$
\end{ex}

\begin{ex}[CVaR]\label{ex:cvar_population_oceform}
A very important special case of OCE risk is CVaR, given earlier in the form (\ref{eqn:cvar_population}) in \S{\ref{sec:core_concepts_riskaverse}}. One of the central results in the seminal work of \citet{rockafellar2000a,rockafellar2002a} is that CVaR as given by (\ref{eqn:cvar_population}) can be represented using a convex program, namely that
\begin{align}\label{eqn:cvar_population_oceform}
\crit_{\ddist}^{\text{CVaR}}(h;\beta) = \inf_{\theta \in \RR}\left[\theta + \frac{1}{1-\beta}\exx_{\ddist}(\loss(h)-\theta)_{+}\right] = \quant_{\beta}(h) + \frac{1}{1-\beta}\exx_{\ddist}(\loss(h)-\quant_{\beta}(h))_{+}
\end{align}
holds for any $0 \leq \beta < 1$, where $(x)_{+} \defeq \max\{0,x\}$ denotes the positive part. In particular, this tells us that CVaR is an OCE risk with $\phi$ set to $\phi(x) = (x)_{+}/(1-\beta)$. In the special case of binary classification using margin errors $\ell(h;x,y) = -y\langle h, x \rangle/\Abs{h}$ with $y \in \{-1,+1\}$, \citet{takeda2008a} showed that minimizing the CVaR objective (\ref{eqn:cvar_population_oceform}) is equivalent to solving the extended $\nu$-SVM objective. On the other hand, when the underlying training losses $\loss_{1}(h),\ldots,\loss_{n}(h)$ are Bernoulli random variables (e.g., 0-1 error), \citet{zhai2021b} show how the CVaR objective collapses into the expected 0-1 error objective, though randomizing the choice of $h$ can be used to obtain distinct performance guarantees.\hfill$\blacksquare$
\end{ex}

\begin{rmk}[Inverted OCE]\label{rmk:inverted_oce}
While traditional OCE risks are always at least as large as the expected loss, giving us flexibility over sensitivity to tails in the upward direction, one naturally may be interested in prioritizing tails in the \emph{downward} direction, i.e., placing more emphasis on ``easy'' examples and ignoring hard examples. Obviously this can be realized directly using order-based risks as discussed in \S{\ref{sec:lcrit_classes_orderbased}}, but a slight modification to the OCE risk class can also be used to achieve such as effect. Letting $\phi$ be as before, define the \term{inverted OCE risk} by
\begin{align}\label{eqn:inverted_oce}
\underline{\text{IOCE}}_{\ddist}(\loss;\phi) & \defeq \sup_{\theta \in \RR} \left[ \theta - \exx_{\ddist}\phi(\theta-\loss)\right]\\
& = -\inf_{\theta \in \RR}\left[ \exx_{\ddist}\phi(\theta-\loss) - \theta \right]\\
& = \exx_{\ddist}\loss - \inf_{\theta \in \RR} \exx_{\ddist}\varphi(\theta-\loss)
\end{align}
where $\varphi(x) = \phi(x) - x$ as before, meaning that since $\varphi(x) \geq 0$ for all $x \in \RR$, the inverted OCE can be no greater than the mean, i.e., $\underline{\text{IOCE}}_{\ddist}(\loss;\phi) \leq \exx_{\ddist}\loss$ holds for all valid $\phi$. The naming is due to \citet{lee2020a}, who also show that the special case of ``inverted CVaR'' under the empirical distribution yields an ``average bottom-$k$'' risk \citep[Prop.~1]{lee2020a}, a fact analogous to the link seen earlier between traditional CVaR and average top-$k$ risk (cf.~\S{\ref{sec:core_concepts_riskaverse}} and Ex.~\ref{ex:average_topk}). In particular, this shows us that inverted OCE can be used to completely ignore outliers in the upward direction; of course, when $\loss$ is unbounded below, sensitivity to outliers in the downward direction is amplified. Similarly, recalling the special case of the tilted/entropic risk introduced earlier with $\phi_{\gamma}(x) = (\mathrm{e}^{\gamma{x}}-1)/\gamma$ for $\gamma > 0$, plugging this in to the inverted OCE (\ref{eqn:inverted_oce}), a bit of calculus quickly shows us that
\begin{align*}
\underline{\text{IOCE}}_{\ddist}(\loss;\phi_{\gamma}) = \frac{1}{-\gamma} \log\left(\exx_{\ddist}\mathrm{e}^{-\gamma\loss}\right)
\end{align*}
meaning that the tilted risk (\ref{eqn:tilted_risk}) using a \emph{negative} tilting parameter is an inverted OCE risk, as we might intuitively expect. Note that in general, the inverted OCE does not inherit the convexity properties enjoyed by the original OCE.\hfill$\blacksquare$
\end{rmk}

\begin{rmk}[Dispersion and OCE]\label{rmk:dispersion_and_oce}
In light of the alternate form (\ref{eqn:oce_risk_varphi}) for OCE risk, one natural entry point when designing new learning criteria is to start with $\varphi$ used to measure ``dispersion'' in the underlying distribution. The canonical measure of dispersion is the standard deviation, and noting that
\begin{align*}
\inf_{\theta \in \RR} \exx_{\ddist}(\loss-\theta)^{2} = \exx_{\ddist}(\loss-\exx_{\ddist}\loss)^{2} = \vaa_{\ddist}\loss
\end{align*}
it seems intuitive to try setting $\varphi(x) = x^{2}/2$. In this case, $\phi(x) = \varphi(x)+x$ indeed satisfies $1 \in \partial\phi(0)$ and $\phi(0)=0$, but a lack of monotonicity means this choice, namely the mean-variance $\exx_{\ddist}\loss + \vaa_{\ddist}\loss$, is not a valid OCE risk.\footnote{Table 1 of \citet{lee2020a} incorrectly includes the mean-variance as an OCE risk.} Naively replacing this with $\phi(x) = (x)_{+}^{2}/2 + x$ does yield a valid OCE risk, but unfortunately it is not very interesting, since
\begin{align*}
\inf_{\theta \in \RR} \exx_{\ddist}(\loss-\theta)_{+}^{2} = 0
\end{align*}
and thus the OCE risk collapses into the mean, i.e., $\text{OCE}_{\ddist}(\loss;\phi) = \exx_{\ddist}\loss$. This also means that criteria based on asymmetric measures of dispersion relative to a fixed threshold $\theta$, such as
\begin{align}\label{eqn:oce_risk_varphi_semidev_p}
\exx_{\ddist}\loss + \left(\frac{1}{p}\right)\exx_{\ddist}\left(\loss - \theta\right)_{+}^{p}
\end{align}
and
\begin{align}\label{eqn:oce_risk_varphi_semidev}
\exx_{\ddist}\loss + \left(\exx_{\ddist}\left(\loss - \theta\right)_{+}^{p}\right)^{1/p}
\end{align}
cannot be OCE risks for any value of $p \geq 1$ and $\theta \in \RR$ such that $\exx_{\ddist}\left(\loss - \theta\right)_{+}^{p} > 0$. In the special case where we set $\theta = \exx_{\ddist}\loss$, the criterion (\ref{eqn:oce_risk_varphi_semidev}) is known as \term{mean-upper-semi-deviation}, and is well-studied in the literature, but not an OCE risk; see the examples in \citet{ruszczynski2006b,rockafellar2013a} for more background.\hfill$\blacksquare$
\end{rmk}

\subsection{DRO Risks}\label{sec:lcrit_classes_dro}

To motivate the next class of learning criteria, let us re-consider ``worst case scenarios'' in a learning problem. Given a fixed data distribution $\rdv{Z} \sim \ddist$ and random test loss $\loss \in \LL_{\ddist}(\HH)$ with $\LL_{\ddist}(\HH)$ as defined in (\ref{eqn:loss_distro_set}) as usual, in the preceding \S{\ref{sec:lcrit_classes_oce}} we saw OCE risks such as entropic risk and CVaR which let us control the degree of sensitivity to the loss tails on the upside, i.e., a bad event stated in terms of the random draw of $\loss$. This scenario is perfectly natural, but recalling the key assumption (\ref{eqn:iid_training_data}) it usually assumes that we have training data $\Z_{n} = (\rdv{Z}_{1},\ldots,\rdv{Z}_{n})$ which are independently sampled from $\ddist$, and does not consider the possibility that the data distribution at training time might differ from that at test time; a case of \term{distribution shift}. In order to make a decision with some degree of resilience to a shift in the data distribution, letting $\PP$ denote a set of probability distributions defined on the same domain as $\ddist$, one can formalize a new ``worst-case'' scenario using
\begin{align}\label{eqn:dro_risk_general}
\overbar{\crit}_{\PP}(h) \defeq \sup\left\{\crit_{\nu}(h): \nu \in \PP\right\}
\end{align}
where $\loss \mapsto \crit_{\nu}(\loss)$ is an arbitrary learning criterion defined on $\LL_{\nu}(\HH)$ for each $\nu \in \PP$, and we overload our notation with $\crit_{\nu}(h) \defeq \crit_{\nu}(\loss(h))$ as usual. In the special case where $\crit_{\nu}\loss = \exx_{\nu}\loss$, the objective (\ref{eqn:dro_risk_general}) and algorithms designed to minimize it are categorized under the term \term{distributionally robust optimization (DRO)}, with (\ref{eqn:dro_risk_general}) referred to as a \term{DRO risk}. See \citet{shapiro2017a} and the references therein for more background from the optimization community. In the context of machine learning, the work of \citet{duchi2021a} has been particularly influential.

It goes without saying that the critical element of (\ref{eqn:dro_risk_general}) is the set $\PP$, often referred to as the \term{uncertainty set} \citep{duchi2021a,shapiro2017a}. Still assuming the training data are sampled from $\ddist$, there must be some connection between $\ddist$ and the elements of $\PP$ in order for the learning problem to be tractable. Even when $\ddist \in \PP$, if all the distributions in $\PP \setminus \{\ddist\}$ are completely unrelated to $\ddist$, then we have no hope of minimizing $\overbar{\crit}_{\PP}(h)$ in (\ref{eqn:dro_risk_general}), even approximately. We only have access to a sample $\Z_{n}$ from $\ddist$, and so the other elements in $\PP$ must be ``similar'' or ``close'' enough to $\ddist$ that a straightforward perturbation of the sample $\Z_{n}$ can effectively simulate a sample taken from the most ``unlucky'' distribution included in $\PP$.

A natural way of formalizing this notion is to introduce a measure of discrepancy between probability distributions. Letting $\Div(\cdot)$ be a function which takes a pair of probability measures and returns a non-negative value from the extended real line, given the special role of $\ddist$ here, it is intuitive to constrain $\PP$ in (\ref{eqn:dro_risk_general}) to contain only distributions which are $\varepsilon$-similar to $\ddist$, where discrepancy is measured by $\Div(\cdot)$. More explicitly, we set $\PP = \PP_{\varepsilon}(\ddist;\Div)$ defined by
\begin{align}\label{eqn:dro_uncertainty_divergence}
\PP_{\varepsilon}(\ddist;\Div) \defeq \left\{ \nu: \Div(\nu;\ddist) \leq \varepsilon \right\}.
\end{align}
If in addition to $\Div(\nu;\ddist) \geq 0$, we have $\Div(\nu;\ddist) = 0$ if and only if $\nu = \ddist$, it is traditional to call $\Div(\cdot)$ a \term{divergence}.\footnote{There is a tremendous literature on divergences and their role in the problems of statistics, machine learning, and information theory. See \citet{grunwald2004a} for a stimulating entry point.} The best-known special case of a divergence is the \term{Kullback-Leibler (KL) divergence} of $\nu$ from $\ddist$, defined by
\begin{align}\label{eqn:kl_divergence}
\Div_{\text{KL}}(\nu;\ddist) \defeq \exx_{\nu}\log\left( \frac{\dif{\nu}}{\dif{\ddist}} \right)
\end{align}
whenever $\nu \ll \ddist$ (absolute continuity), with $\Div_{\text{KL}}(\nu;\ddist) \defeq \infty$ otherwise. Here $\dif{\nu}/\dif{\ddist}$ denotes the Radon-Nikodym density of $\nu$ with respect to $\ddist$, a random variable.\footnote{For a textbook introduction to Radon-Nikodym densities, see \citet[\S{2.2}]{ash2000a}.} Note that $\Div_{\text{KL}}$ in (\ref{eqn:kl_divergence}) can be re-written as
\begin{align}\label{eqn:kl_divergence_newform}
\Div_{\text{KL}}(\nu;\ddist) = \exx_{\ddist}\left( \frac{\dif{\nu}}{\dif{\ddist}} \right)\log\left( \frac{\dif{\nu}}{\dif{\ddist}} \right) = \exx_{\ddist}\left( \frac{\dif{\nu}}{\dif{\ddist}} \right)\log\left( \frac{\dif{\nu}}{\dif{\ddist}} \right) - \exx_{\ddist}\left( \frac{\dif{\nu}}{\dif{\ddist}} \right) + 1.
\end{align}
In the right-most expression of (\ref{eqn:kl_divergence_newform}), the density $\dif{\nu}/\dif{\ddist}$ is being passed through the function $x \mapsto x\log(x) - x + 1$ before taking expectation. This function is convex and non-negative on $[0,\infty)$, taking its minimum (zero) at $x=1$. An important generalization of the KL divergence simply replaces this map with an arbitrary function $x \mapsto f(x)$ that retains the key properties just mentioned. More formally, given a convex function $f:[0,\infty) \to [0,\infty]$ satisfying $f(1)=0$, we define the \term{$f$-divergence} of $\nu$ from $\ddist$ by
\begin{align}\label{eqn:f_divergence}
\Div_{f}(\nu;\ddist) \defeq \exx_{\ddist}f\left(\frac{\dif{\nu}}{\dif{\ddist}}\right)
\end{align}
whenever $\nu \ll \ddist$, otherwise setting $\Div_{f}(\nu;\ddist) \defeq \infty$ as before. While in principle all sorts of different uncertainty sets can be conceived of, typical DRO risks in the machine learning literature are all special cases of $\PP_{\varepsilon}(\ddist;\Div_{f})$. See recent work by \citet{frohlich2023a}, who demonstrate sharp connections between the design of $\PP_{\varepsilon}(\ddist;\Div_{f})$ and distributions with ``desirable'' tail properties. Representative special cases of DRO risk will be introduced in the examples to follow.

\begin{ex}[KL, DRO, and tilted risk]\label{ex:dro_kl_tilted}
A natural first example is $f(x) = x\log(x) - x + 1$, so that $\Div_{f} = \Div_{\text{KL}}$ as given above. In this case, assuming that $\crit_{\nu}(h) = \exx_{\nu}\loss(h)$ and the absolute loss $\abs{\loss(h)}$ is bounded, when we set $\PP = \PP_{\varepsilon}(\ddist;\Div_{\text{KL}})$ for $\varepsilon > 0$, the DRO risk (\ref{eqn:dro_risk_general}) can be expressed as
\begin{align}\label{eqn:dro_kl_tilted}
\overbar{\crit}_{\PP}(h) = \inf_{\gamma > 0} \left[ \frac{\varepsilon}{\gamma} + \frac{1}{\gamma}\log\left(\exx_{\ddist}\mathrm{e}^{\gamma\loss(h)} \right) \right].
\end{align}
See \citet[Ex.~3.11]{shapiro2017a} and the references therein for more details on this derivation. The equality (\ref{eqn:dro_kl_tilted}) illustrates a close relation between the tilted risk under $\ddist$ (cf.~Example \ref{ex:tilted_risk}) and the worst-case \emph{mean} over all $\nu$ which are at least $\varepsilon$-similar to $\ddist$ in terms of KL divergence. Taking $\varepsilon > 0$ large means we need to take $\gamma > 0$ correspondingly large in order to minimize the objective in (\ref{eqn:dro_kl_tilted}). Taking $\varepsilon \to 0_{+}$ implies that the optimal tilting parameter correspondingly decreases as $\gamma \to 0_{+}$. In this limit we have $\overbar{\crit}_{\PP}(h) \to \exx_{\ddist}\loss(h)$. Note however that minimizing (\ref{eqn:dro_kl_tilted}) in $h$ is not in general equivalent to minimizing the $\gamma$-tilted risk (\ref{eqn:tilted_risk}) for any fixed choice of $\gamma$, since the optimal $\gamma$ in (\ref{eqn:dro_kl_tilted}) changes with the choice of $h$.\hfill$\blacksquare$
\end{ex}

\begin{ex}[Cressie-Read DRO]
A flexible and useful sub-class of $f$-divergences is given by setting $f=f_{c}$ for functions of the form
\begin{align}\label{eqn:cressie_read}
f_{c}(x) \defeq \frac{x^{c}-cx+c-1}{c(c-1)}
\end{align}
parameterized by any $c \in \RR \setminus \{0,1\}$. In the limit of $c \to 1$, we have $f_{c}(x) \to x\log(x) - x + 1$ and thus we recover KL divergence as a special case. This class is called the \term{Cressie-Read family} of $f$-divergences by \citet{duchi2021a}, and restricting ourselves to $c > 1$, setting $\PP = \PP_{\varepsilon}(\ddist;\Div_{f_{c}})$ we have
\begin{align}\label{eqn:dro_cressie_read}
\overbar{\crit}_{\PP}(h) = \inf_{\theta \in \RR} \left[ \theta + (1+c(c-1)\varepsilon)^{1/c} \left( \exx_{\ddist}(\loss(h)-\theta)_{+}^{c_{\ast}} \right)^{1/c_{\ast}} \right]
\end{align}
for any choice of $\varepsilon > 0$ \citep[Lem.~1]{duchi2021a}, where $c_{\ast} \defeq c/(c-1)$. The expression in (\ref{eqn:dro_cressie_read}) is quite reminiscent of the OCE risk classes seen earlier, and indeed the basic effect is very similar. Setting $\phi_{c,\varepsilon}(x) \defeq (1+c(c-1)\varepsilon)^{c_{\ast}/c}(x)_{+}^{c_{\ast}}$, we have
\begin{align}\label{eqn:dro_cressie_read_simplified}
\overbar{\crit}_{\PP}(h) = \inf_{\theta \in \RR} \left[ \theta + \left( \exx_{\ddist}\phi_{c,\varepsilon}(\loss(h)-\theta) \right)^{1/c_{\ast}} \right]
\end{align}
with $\phi_{c,\varepsilon}$ satisfying the key monotonicity and convexity properties required when making OCE risks; the only difference is the $(\cdot)^{1/c_{\ast}}$ wrapping the expected value. In the end, both Cressie-Read DRO risks and OCE risks give us control over how much priority to put on ``difficult'' examples, where higher difficulty corresponds to larger losses incurred by a given $h \in \HH$. Both classes are also risk-averse in the sense that they always upper-bound the expected loss $\exx_{\ddist}\loss(h)$. High tail sensitivity means that empirical estimates will have to deal with outliers, making efficient estimation and learning under finite samples a challenge; as a natural compromise, \citet{zhai2021a} propose a variant of DRO risks which emphasizes values of sufficient difficulty, but ignores everything beyond that threshold.\hfill$\blacksquare$
\end{ex}

\begin{ex}[Variance-based regularization]
DRO risks received attention early on in the context of reducing both the expected loss and its variance in order to obtain faster convergence rate guarantees. Key initial works are the pre-prints \citep{duchi2016b,duchi2016a} and the conference paper of \citet{namkoong2016a}, later distilled into the journal papers \citep{duchi2019a,duchi2021b}. Of particular importance is the fact the (empirical) minimizer of DRO risk (\ref{eqn:dro_risk_general}) using an appropriate choice of $f$-divergence (e.g., $\chi^{2}$-divergence) can be shown to enjoy high-probability bounds on the expected loss taking the form
\begin{align*}
\inf_{h \in \HH}\left[ \exx_{\ddist}\loss(h) + \bigO\left(\sqrt{\frac{\vaa_{\ddist}\loss(h)}{n}}\right) \right] + \bigO\left(\frac{1}{n}\right)
\end{align*}
where $\bigO(\cdot)$ hides constants and factors depending on the confidence level for simplicity; see \citet{duchi2019a} and the references therein for more details.\hfill$\blacksquare$
\end{ex}

\begin{ex}[Sub-population awareness]
A well-explored application of DRO risks is the design of learning algorithms which have comparable performance across certain ``sub-populations.'' Formally, if $\rdv{Z} \sim \ddist$ is our random data, we can formulate $k$ sub-populations as sets $Z_{1},\ldots,Z_{k}$ that $\rdv{Z}$ may fall into. For some intuitive examples, these sets could represent data belonging to distinct ethnic, age, or gender categories, where $k$ is the number of categories. Recalling \S{\ref{sec:core_concepts_fairness}} and the idea of fairness-aware learning, a natural goal is to reduce the gap in performance across sub-populations that tends to arise under naive optimization of the expected loss; this is called ``representation disparity'' by \citet{hashimoto2018a}. One direct way of doing this is to consider a worst-case objective over sub-populations, such as
\begin{align}\label{eqn:risk_worst_subpop}
\max_{j \in [k]} \exx_{\ddist}\left[ \loss(h) \cond \rdv{Z} \in Z_{j}\right].
\end{align}
The objective in (\ref{eqn:risk_worst_subpop}) can be bounded using DRO risk in a straightforward manner. Letting $\nu_{j}$ denote the conditional distribution of $\rdv{Z} \cond \rdv{Z} \in Z_{j}$ for each $j \in [k]$, clearly for any $Z$ such that $\ddist\{\rdv{Z} \in Z\} = 0$, we have $\nu_{j}\{\rdv{Z} \in Z\} = 0$, meaning $\nu_{j} \ll \ddist$ holds, and we can find a constant $0 < r_{\text{max}} < \infty$ such that
\begin{align}
\max_{j \in [k]} \frac{\dif{\nu_{j}}}{\dif{\ddist}} \leq r_{\text{max}}
\end{align}
with probability 1. It thus follows that
\begin{align}
\Div_{f}(\nu_{j};\ddist) = \exx_{\ddist}f\left(\frac{\dif{\nu_{j}}}{\dif{\ddist}}\right) \leq f\left(r_{\text{max}}\right)
\end{align}
and thus setting $\varepsilon_{\text{max}} \defeq f\left(r_{\text{max}}\right)$, we have $\nu_{j} \in \PP_{\varepsilon_{\text{max}}}(\ddist;\Div_{f})$ for each $j \in [k]$. We may thus conclude that
\begin{align}
\max_{j \in [k]} \exx_{\ddist}\left[ \loss(h) \cond \rdv{Z} \in Z_{j}\right] = \max_{j \in [k]} \exx_{\nu_{j}}\loss(h) \leq \overbar{\crit}_{\PP}(h)
\end{align}
with $\PP = \PP_{\varepsilon_{\text{max}}}(\ddist;\Div_{f})$ as above. This useful and general fact is stated concisely by \citet[Prop.~1]{zhai2021a}, though it appears earlier for the special case of $f=f_{c}$ following (\ref{eqn:cressie_read}) with $c=2$ (known as $\chi^{2}$-divergence) in \citet[Prop.~2]{hashimoto2018a}. This fact is useful since it suggests that minimizing the DRO risk may be a fruitful strategy even in the ``oblivious'' setting where the sub-populations are unknown.\hfill$\blacksquare$
\end{ex}

\subsection{Miscellaneous}

In the preceding \S{\ref{sec:lcrit_classes_orderbased}}--\S{\ref{sec:lcrit_classes_dro}} we covered the three best-known classes of learning criteria in the machine learning literature. In this section, we briefly introduce some other classes which have appeared more recently in the literature.

\subsubsection{Criteria Based on Bidirectional Dispersion}

In our description earlier of OCE and DRO risks (with uncertainty set as in (\ref{eqn:dro_uncertainty_divergence})), we saw how these risks can be used to emphasize the tails in the upward direction, meaning that for any criterion $\crit_{\ddist}$ from either of these classes, we have
\begin{align}\label{eqn:bdd_onedirectional_averse}
\exx_{\ddist}\loss \leq \crit_{\ddist}(\loss), \quad \forall \, \loss \in \LL_{\ddist}(\HH)
\end{align}
regardless of the properties of $\ddist$ and $\HH$. We also saw an ``inverted'' version of OCE risks (cf.~Remark \ref{rmk:inverted_oce}) which can be used to emphasize the tails on the downside, resulting in
\begin{align}\label{eqn:bdd_onedirectional_seeking}
\exx_{\ddist}\loss \geq \crit_{\ddist}(\loss), \quad \forall \, \loss \in \LL_{\ddist}(\HH).
\end{align}
In any case, all OCE risks (inverted variants included) and DRO risks are ``uni-directional'' in the sense that either (\ref{eqn:bdd_onedirectional_averse}) or (\ref{eqn:bdd_onedirectional_seeking}) holds; these criteria can only be ``pulled'' by heavy tails in one direction, never two. More precisely, we cannot have two distinct distributions $\loss,\loss^{\prime} \in \LL_{\ddist}(\HH)$ where $\exx_{\ddist}\loss \geq \crit_{\ddist}(\loss)$ and $\exx_{\ddist}\loss^{\prime} < \crit_{\ddist}(\loss^{\prime})$ hold simultaneously. Of course, certain order-based criteria readily satisfy this. Considering the median for example, if $\quant_{0.5}(\loss) > \exx_{\ddist}\loss$, then trivially $\quant_{0.5}(-\loss) < \exx_{\ddist}(-\loss)$. On the other hand, as mentioned in \S{\ref{sec:core_concepts_robustness}}, while quantiles, like any M-parameter, are easy to compute, they are difficult to optimize as a function of the loss distribution. A potential compromise can be achieved using learning criteria of the general form
\begin{align}\label{eqn:trisk_general}
\crit_{\ddist}(\loss;\theta,\eta,\rho) \defeq \eta\theta + \exx_{\ddist}\rho\left(\loss-\theta\right)
\end{align}
where $\eta \in \RR$ (negative values are allowed), $\theta \in \RR$, and $\rho:\RR \to [0,\infty)$ is assumed to be such that the induced set of M-parameters is non-empty and satisfies $\argmin_{\theta \in \RR} \exx_{\ddist}\rho\left(\loss-\theta\right) \subset \RR$. The criterion $\crit_{\ddist}(\loss;\theta,\eta,\rho)$ is called the \term{threshold risk} (or \term{T-risk}) by \citet{holland2023bdd}. A natural sub-class is the ``minimal T-risk'' given by
\begin{align}\label{eqn:trisk_minimal}
\underline{\crit}_{\ddist}(\loss;\eta,\rho) \defeq \inf_{\theta \in \RR} \crit_{\ddist}(\loss;\theta,\eta,\rho)
\end{align}
which requires that $\rho$ grow without bound at a linear rate (or faster); otherwise trivially $\underline{\crit}_{\ddist}(\loss;\eta,\rho) = -\infty$. This is a compromise in the sense that the criterion itself never yields the M-parameters induced by $\rho$ exactly, only a ``regularized'' version biased slightly above or below, depending on $\sign(\eta)$. On the other hand, it is computationally much easier to handle than true M-parameters, with minimization of $\crit_{\ddist}(\loss;\theta,\eta,\rho)$ in $(\loss,\theta)$ quite easy to implement. The minimal T-risk (\ref{eqn:trisk_minimal}) has obvious similarities to OCE risks from \S{\ref{sec:lcrit_classes_oce}}, but the main difference is that $\rho(\cdot)$ can be \emph{non-monotonic}. This means that convexity guarantees for $h \mapsto \crit_{\ddist}(\loss(h);\theta,\eta,\rho)$ are not in general possible, but it gives us the flexibility to set $\eta$ to both positive and negative values for a fixed $\rho$, and allows for bidirectional sensitivity to loss tails that is not possible with OCE or DRO risks.\footnote{Early work on this class for a rather specialized choice of $\rho$ dates back to a 2020 pre-print by \citet{holland2020mrisk}, later expanded by \citet{holland2022c}. See \citet{holland2023bdd} for more general results.} This positions the minimal T-risk in a somewhat peculiar middle-ground; it does not appear in the survey of \citet{ruszczynski2006b}, nor on any corner of the ``quadrangles'' in the rich examples of \citet{rockafellar2013a}.

\begin{ex}[Quantile-based thresholds]
Setting $\rho(\cdot)=\abs{\cdot}$ and manipulating $\eta \in (-1,+1)$, we can set the T-risk threshold to any quantile of $\loss$. More precisely, if we set $\eta = 1-2\beta$ for any quantile level $0 < \beta < 1$, we have
\begin{align*}
\underline{\crit}_{\ddist}(\loss;\eta,\rho) = (1-2\beta)\quant_{\beta}(\loss) + \exx_{\ddist}\rho\abs{\loss-\quant_{\beta}(\loss)}.
\end{align*}
For more on properties defined using convex programs, see \citet{koltchinskii1997a}.\hfill$\blacksquare$
\end{ex}

\begin{ex}[Weakly convex T-risk]
While the objective $h \mapsto \crit_{\ddist}(\loss(h);\theta,\eta,\rho)$ will not in general be convex when $\rho$ is non-monotonic, a form of weak convexity can be guaranteed when the losses are sufficiently smooth (in $h$) using convex $\rho$ with sub-quadratic growth in the limit. One useful example is
\begin{align}
\rho(x) = x\atan(x) - \frac{\log(1+x^{2})}{2}
\end{align}
whose strict convexity and Lipschitz continuity are amenable to weak convexity arguments, and coupled with a re-scaling mechanism lets us flexibly interpolate the minimizing threshold between $\exx_{\ddist}\loss$ and $\quant_{0.5}(\loss)$ \citep{holland2022c}.\hfill$\blacksquare$
\end{ex}

\subsubsection{Criteria Inspired by Human Psychology}

A unique and interesting direction is designing learning criteria which attempt to mimic human tendencies for risk aversion (or preference) when making decisions under uncertainty. In terms of the machine learning literature, the most salient example is an application of the metrics used in the \term{cumulative prospect theory (CPT)} of \citet{tversky1992a}. Roughly speaking, ``prospect'' refers to subjective evaluations of uncertain outcomes. To make our explanation as clear as possible, let us denote this uncertain outcome using an arbitrary random variable $\rdv{X} \sim \ddist$. Let $\valfn(\cdot)$ be a ``value'' function, and let $\dfun_{\ddist}$ denote the distribution function of $\rdv{X}$, namely $\dfun_{\ddist}(x) \defeq \prr\{\rdv{X} \leq x\}$. The \term{CPT score} assigned to $\rdv{X}$ is defined by taking the expectation using a \emph{modified} distribution function $\widetilde{\dfun}_{\ddist}$, namely
\begin{align}
\cpt_{\ddist}(\rdv{X}) \defeq \int_{\RR} \valfn(x) \, \widetilde{\dfun}_{\ddist}(\dif x).
\end{align}
Thus, the CPT score is characterized by the value function $\valfn$ and the transformation $\dfun_{\ddist} \mapsto \widetilde{\dfun}_{\ddist}$. Setting $c$ as a threshold (often $c=0$), using indicators $\indic\{\rdv{X} \geq c\}$ and $\indic\{\rdv{X} < c\}$ the CPT score typically treats these two mutually exclusive events differently. To do this, we construct value functions $\valfn_{+}$ and $\valfn_{-}$ and differentiable weighting functions $\myw_{+}$ and $\myw_{-}$ for each of these cases, setting
\begin{align*}
\valfn(x) & = \indic\{ x \geq c \}\valfn_{+}(x) + \indic\{ x < c \}\valfn_{-}(x)\\
\widetilde{\dfun}(x) & = \indic\{ x \geq c \}\myw_{+}(\dfun(x)) + \indic\{ x < c \}\myw_{-}(\dfun(x)).
\end{align*}
Putting these pieces together, the resulting CPT score takes the more explicit form
\begin{align}
\cpt_{\ddist}(\rdv{X}) = \int_{-\infty}^{c} \valfn_{-}(x) \myw_{-}^{\prime}(\dfun(x)) \, \dfun(\dif x) + \int_{c}^{\infty} \valfn_{+}(x) \myw_{+}^{\prime}(\dfun(x)) \, \dfun(\dif x)
\end{align}
where $\myw_{+}^{\prime}$ and $\myw_{-}^{\prime}$ are the derivatives of the weight functions. The threshold of $c=0$ is often used, but this is of course arbitrary; it can be set to any meaningful value given the context. The important concepts here are as follows: (a) ``gains'' and ``losses'' relative to some threshold are treated differently in human psychology; (b) small probabilities tend to be underweighted, large probabilities tend to be overweighted. See \citet{rieger2006a} for a useful background reference. Taking this from CPT scores to a natural notion of \term{CPT risk} just amounts to setting $\rdv{X} = \loss(h)$, with our usual shorthand notation $\cpt(h) \defeq \cpt(\loss(h))$ for each $h \in \HH$. While the general-purpose CPT score itself is well-known, literature on the CPT risk in machine learning remains quite sparse. Concentration bounds for estimators of the CPT risk (pointwise in $h$) are given by \citet{bhat2019a}. To the best of our knowledge, it appears that \citet{liu2019a} are the first to consider learning with CPT risks, under the catchy name ``human-aligned risks,'' although their studies are limited to an empirical analysis of learning using ERM implementations of certain special sub-classes of CPT risks. Under some assumptions, statistical learning guarantees are possible for specialized sub-classes of CPT risk using the machinery of \citet{liu2022a}.

\begin{ex}[Links to order-based risks]
Recalling our formulation of order-based risks in \S{\ref{sec:lcrit_classes_orderbased}}, there is clear conceptual overlap with the CPT risk just introduced. Using a weight function
\begin{align*}
W_{\text{CPT}}(u) \defeq \indic\left\{u \geq \dfun_{\ddist}(c)\right\}\myw_{+}^{\prime}(u) + \indic\{ u < \dfun_{\ddist}(c) \}\myw_{-}^{\prime}(u)
\end{align*}
we see that the criterion $\crit_{\ddist}(\loss;W)$ in (\ref{eqn:quantile_risk_pop_expform}) can be used to represent a particular CPT risk, i.e., leaving the value function as the identity $\valfn(x) = x$, setting $W=W_{\text{CPT}}$ we have
\begin{align*}
\cpt_{\ddist}(\loss) = \crit_{\ddist}(\loss;W_{\text{CPT}}).
\end{align*}
The general value function is not present in our formulation in \S{\ref{sec:lcrit_classes_orderbased}}, but it should be noted that arbitrary (value) functions of order statistics are reflected in the textbook formulation of L-estimates given by \citet[Ch.~3]{huber2009a}.\hfill$\blacksquare$
\end{ex}

\section{Historical Background}\label{sec:history}

To complement the preceding survey of criteria used in machine learning, we take a deeper look at the historical context within which the ``classical paradigm'' described in \S{\ref{sec:intro_classical}} developed.

\subsection{Origins of Expected Loss Minimization}\label{sec:history_expectation}

The problem of \emph{decision-making under uncertainty} has interested scholars for centuries. For an uncertain outcome which can be expressed numerically, the traditional ``mathematical expectation'' is the sum of all possible outcomes, weighted by the probability of each outcome.\footnote{Historically, a broader notion of ``expectation'' can in many ways be seen as a more basic (and intuitive) concept than that of probability. Expectations are central to fair pricing in games of chance described by Christian Huygens in 1657, see \citet[Ch.~11]{hacking2006ProbHistory} for detailed background, and \citet[Ch.~1]{pollard2002UGMTP} for similar ideas in the context of measure theoretic foundations.} This notion was clearly described in letters from the Bernoullis and Gabriel Cramer dating back to the early 18th century \citep{bassett1987a}. With the expected value defined as a probability-weighted sum, then a problem of fundamental importance is \emph{how to evaluate outcomes}, namely the process of assigning numerical values to real-world outcomes, which are appropriate within the context of the decisions that need to be made. The importance of this is highlighted by the ``St.~Petersburg paradox,'' dating back to Nikolaus and Daniel Bernoulli \citep{menger1967a}. Consider the following game: a fair coin is tossed until it turns up heads, and one receives a payment of $2^{\rdv{K}}$ dollars (or ducats), where $\rdv{K}$ is a random variable representing the number of tosses it took for heads to turn up. Note that $\rdv{K}$ is bounded below by $1$, and unbounded above. The expected payout is $\exx[2^{\rdv{K}}] = 2(1/2) + 2^{2}(1/2^{2}) + \cdots = \infty$. While the expected payout is infinite, it seems irrational to pay a large amount to play this game, and thus arises the need to refine how we evaluate outcomes when making decisions. Essentially, this amounts to transforming the raw payouts in such a way that better reflects human preferences, say by introducing a function $f$ such that $\exx[f(2^{\rdv{K}})] < \infty$. More generally, given any uncertain outcome $\rdv{X}$, the transformed value $f(\rdv{X})$ is often called the \term{utility} of that outcome.\footnote{The expected utility was historically referred to as the ``moral expectation'' (in contrast to the mathematical expectation) by Daniel Bernoulli, who suggested using the utility function $f(x) = \log(x)$ as a solution to the St.~Petersburg problem \citep{jordan1924a}. From our modern viewpoint, it is clear that this only solves one special case of the problem, since given any $f$ which is unbounded above, one can construct an analogous payout scheme such that the expected utility is infinite. The first work known to have pointed out this problem was a 1934 work in German by Karl Menger \citep{menger1967a}; see \citet{bassett1987a} for an interesting discussion on the history of this issue.} The basic decision-making principle derived from this line of thought is simple: act in a way that \emph{maximizes the expected utility}.\footnote{For more history on expected utility and its role in the theory of the social sciences (especially economics), see the works of \citet{jordan1924a}, \citet{uzawa1956a}, \citet{samuelson1977a}, and \citet{bassett1987a}.}

Over the course of the 20th century, the theoretical foundations of this principle were substantially strengthened. Of particular historical significance is the seminal theory of games developed by \citet{vonneumann1953Games}, where expected utility $\exx[f(\rdv{X})]$ appears in the context of axioms related to preferences over uncertain outcomes (or ``lotteries'') that we expect any ``rational'' agent to satisfy. These axioms are typically stated with respect to a class of uncertain outcomes, here a set $\XX$ of random variables. The famous ``expected utility theorem'' states that any agent satisfying these axioms (under $\XX$) \emph{must} assign its preferences based on expected utility, that is, there exists a utility function $f$ such that for any $\rdv{X}_{1}, \rdv{X}_{2} \in \XX$, we have $\rdv{X}_{1} \preceq \rdv{X}_{2}$ (i.e., ``$\rdv{X}_{2}$ is no worse than $\rdv{X}_{1}$'') if and only if $\exx[f(\rdv{X}_{1})] \leq \exx[f(\rdv{X}_{2})]$.\footnote{For a modern proof of this fact for discrete random variables, see \citet[Props.~6.B.3, 6.E.1]{mascolell1995MicroEcon}. Regarding the terminology used by von Neumann and Morgenstern, they use the term ``utility'' to describe the real-world concept to which we attach preferences, and ``numerical utility'' to refer to functions that satisfy the properties of what we call the expected utility \citep[Sec.~3.5.1]{vonneumann1953Games}. The requirements that they place on numerical utilities (e.g., their axioms (3:1:a) and (3:1:b)) can be interpreted as conditions on maps from $\XX$ to $\RR$, without explicitly involving a function $f$. Some authors refer to $f$ as a \term{Bernoulli utility function}, and $\exx[f(\rdv{X})]$ as the \term{von Neumann-Morgenstern utility} \citep[Ch.~6, footnote 12]{mascolell1995MicroEcon}.} As such, when the agent can choose between uncertain outcomes, it will choose the one with the largest expected utility. Due to the significant influence of this work, the decision-making principle of maximizing the expected utility is essentially synonymous with the ``rational agent'' model pervasive in the social sciences \citep{tversky1992a,mascolell1995MicroEcon}.

Since the usual form of expected utility is a probability-weighted sum, one naturally runs into the question of where these probabilities come from, and what they mean.\footnote{Another important question is that of how to best represent probability in a mathematical form. See \citet[Ch.~1]{pollard2002UGMTP} and the references therein for a start on how the modern answer to this question was developed.} When betting on the outcome of a coin toss, under most conditions it is reasonable to say that the equal probability of heads and tails is a fact of nature, and the probabilities $\prr\{\rdv{X} = \text{``heads''}\} = \prr\{\rdv{X} = \text{``tails''}\} = 1/2$ are in this sense \emph{objectively} determined. On the other hand, consider the question of ``will the stock price of Company ABCD rise next quarter?'' This is a typical example of uncertainty in the real world, and while the possible outcomes are crystal-clear (the price will either rise, or it will not), it is very difficult to conceive of any kind of objective probability that can be assigned to such an uncertain outcome. Depending on knowledge and past experience (among countless other factors), the probability that different parties assign to this event will assuredly differ, making probability an inherently \emph{subjective} notion. Indeed, if we polled a handful of stock traders, employees of Company ABCD, and individuals with no interest or knowledge of the company, we would not be surprised if all these parties had very different expectations.\footnote{Clearly, many decisions under uncertainty will inevitably involve subjective probabilities. Recent books by \citet{taleb2008BlackSwan} and \citet{kahneman2011TFS} aimed at broad audiences give a lucid discussion of the potential ramifications that this subjectivity has on our interaction with complex systems, in particular those of a socioeconomic nature.}

Historically, this dichotomy of \emph{objective and subjective probability} has received a great deal of attention, and underlies much scholarly activity in many disciplines. For example, objective probability can be seen as underlying the \emph{principle of repeated sampling}, which is central to the statistical methodology of Fisher \citep{young2005StatsBook}. Indeed, it is hard to justify a procedure of taking controlled random samples to estimate probability unless that probability is ``well-defined'' in an objective sense. In contrast, Bayesian statistical methods give the statistician the freedom to reflect opinion and knowledge (or a lack thereof) in the inferential procedure, through a prior distribution over the parameter set, which allows for a subjective interpretation of the probabilistic model.\footnote{For a textbook introduction to statistical inference that effectively describes the development of the field within the context of all the major schools of thought, \citet{young2005StatsBook} is a useful reference.} Given identical data sets, different statisticians can arrive at different conclusions. Arguably the most influential proponent of subjective probability is that of Leonard J.~Savage; in his seminal work \citep{savage1954Foundations}, he builds a theory of probability and statistics which is centered on ``personal'' probability. In the context of Bayesian inference, this approach can be implemented by emphasizing a subjective prior \citep[Sec.~3.6]{young2005StatsBook}. Recalling the expected utility theorem described earlier, Savage proves an analogous theorem, but manages to do it without assuming given probabilities (used in computing an ``objective'' expected utility), and instead shows that the rational agent axioms imply a form of \emph{subjective} expected utility maximization.\footnote{See \citet[Sec.~6.F]{mascolell1995MicroEcon} for a high-level description and useful references.} This view of probability and decision-making is prevalent in modern game theory \citep[Sec.~1.4]{osborne1994GameTheory}, and the subjective expected utility theorem is a foundational result in the broader context of decision theory \citep{sep2020DecisionTheory}.

The final important historical development we should highlight is that of viewing \emph{statistical inference as a decision-making problem}, in contrast with the classical notion of inference being a rather mechanical summary of data \citep[Ch.~1--2]{young2005StatsBook}. Of particular importance is the work of Abraham Wald, which laid the foundations for statistical decision theory \citep{wald1939a,wald1945a,wald1949a}. One key theme is that from the statistician's perspective, different types of errors may naturally have a different importance, or weight. Starting with his seminal 1939 work \citep{wald1939a}, Wald introduces a general-purpose loss function (what he calls a ``weight function'') that quantifies this decision-dependent importance, and captures a wide class of statistical problems by formulating them in terms of \emph{minimizing the expected loss}, or what he calls the ``risk.''\footnote{Let us compare Wald's terminology with ours (cf.~\S{\ref{sec:intro_our_view}}). The ``loss'' of \citet{wald1939a} depends on an unknown true model parameter (denoted by $\theta$ in his notation), so his ``loss'' corresponds more closely to a criterion $\crit_{\ddist}$ here, and his ``risk'' then amounts to $\exx[\crit_{\ddist}(\rdv{H}_{n})]$, with expectation taken over $\Z_{n}$.} Underlying this ``risk,'' of course, is a random loss. With Wald's framework as context, a straightforward generalization beyond typical statistical models and tasks leads us directly to the random test loss $\ell(h;\rdv{Z})$ that appears throughout \S{\ref{sec:intro}}--\S{\ref{sec:lcrit_classes}}, depending on the candidate $h \in \HH$. In particular, note that the class of random losses $\{\ell(h;\rdv{Z}): h \in \HH\}$ is in direct analogy with the class of ``lotteries'' $\XX = \{\rdv{X}_{1}, \rdv{X}_{2},\ldots\}$ discussed earlier in the context of expected utility. Furthermore, an important characteristic of the decision-theoretic approach to statistics is an emphasis upon the optimality of \emph{general inferential procedures}, stated in terms that do not depend on any particular dataset \citep[Ch.~1]{young2005StatsBook}. In short, the qualities of data-driven \emph{algorithms} become an important object of analysis, which makes for a smooth methodological transition to machine learning problems.

\subsection{Expected Loss and Learning Problems}\label{sec:history_learning}

Here we provide some additional justification for our position (cf.~\S{\ref{sec:intro_classical}}) that emphasizing performance on average is the dominant paradigm in machine learning. A natural starting point is the Perceptron of Frank Rosenblatt in the late 1950s \citep{rosenblatt1958a}, and the 1962 proof of the convergence of the Perceptron learning algorithm (here, PLA) for linearly separable data.\footnote{See \citet{novikoff1963a} and \citet[Ch.~11]{minsky1969Perceptrons} for background on this result.} This important result is recognized as having effectively ``started learning theory'' \citep[p.~5]{vapnik1999NSLT}. Making this a bit more concrete, given a dataset of $n$ points $\Z_{n} = (\rdv{Z}_{1},\ldots,\rdv{Z}_{n})$, assumed to be input-output pairs $\rdv{Z}_{i} = (\rdv{X}_{i},\rdv{Y}_{i})$ with binary labels $\rdv{Y}_{i} \in \{-1,+1\}$, and denoting the zero-one training losses as $\loss_{i}(h) \defeq \indic\{ h(\rdv{X}_{i}) \neq \rdv{Y}_{i} \}$ for $i \in [n]$, the convergence theorem tells us that whenever perfect accuracy is possible, then the PLA output $\Z_{n} \mapsto \rdv{H}_{\text{PLA}}$ will achieve it, namely that
\begin{align}\label{eqn:PLA_emp_risk}
\frac{1}{n}\sum_{i=1}^{n}\loss_{i}(\rdv{H}_{\text{PLA}}) = 0.
\end{align}
In the context of learning, the question is then how the PLA \emph{generalizes} to new data. Given an independent test point $\rdv{Z} = (\rdv{X},\rdv{Y})$ and writing $\loss(h) = \indic\{h(\rdv{X}) \neq \rdv{Y}\}$, off-sample misclassification is characterized by $\loss(\rdv{H}_{\text{PLA}})$. Since this is just a Bernoulli random variable, its distribution is characterized by its mean. With the key fact of (\ref{eqn:PLA_emp_risk}) in mind, the question of generalization can then be reduced to (uniform) control of the deviations between ``relative frequencies of events and their probability,'' which is precisely the topic of the seminal 1971 paper of \citet{vapnik1971a}.

The approach of the original ``VC theory'' goes well beyond hyperplanes and PLA. Assuming $\rdv{Z}$ and $\Z_{n}$ are independently sampled from $\ddist$, for any model $\HH$ of binary classifiers, write $\widehat{\rdv{C}}_{n}(h) = (1/n)\sum_{i=1}^{n}\loss_{i}(h)$ and $\crit_{\ddist}(h) = \prr\{h(\rdv{X}) \neq \rdv{Y}\}$ for the empirical and true zero-one risk, respectively, and note that for the output $\rdv{H}_{n}$ of any learning algorithm (not just PLA), we trivially have
\begin{align}\label{eqn:uniform_empirical}
\abs{\widehat{\rdv{C}}_{n}(\rdv{H}_{n}) - \crit_{\ddist}(\rdv{H}_{n})} \leq \sup_{h \in \HH} \abs{\widehat{\rdv{C}}_{n}(h) - \crit_{\ddist}(h)}.
\end{align}
The results of \citep{vapnik1971a} enable us to characterize the convergence properties of the right-hand side of (\ref{eqn:uniform_empirical}) based on properties of the model $\HH$ that are of a combinatorial nature. Having this control is particularly useful for evaluating the performance of empirical risk minimizers, namely any algorithm which minimizes $\widehat{\rdv{C}}_{n}$, such as PLA under linearly separable data. Making this explicit, if we constrain the learning algorithm such that $\rdv{H}_{n}$ minimizes $\widehat{\rdv{C}}_{n}$, since for any fixed reference point $h^{\ast} \in \HH$ we have $\widehat{\rdv{C}}_{n}(\rdv{H}_{n}) \leq \widehat{\rdv{C}}_{n}(h^{\ast})$, it follows that
\begin{align}
\nonumber
\crit_{\ddist}(\rdv{H}_{n}) - \crit_{\ddist}(h^{\ast}) & = \crit_{\ddist}(\rdv{H}_{n}) - \widehat{\rdv{C}}_{n}(\rdv{H}_{n}) + \widehat{\rdv{C}}_{n}(\rdv{H}_{n}) - \widehat{\rdv{C}}_{n}(h^{\ast}) + \widehat{\rdv{C}}_{n}(h^{\ast}) - \crit_{\ddist}(h^{\ast})\\
& \leq 2 \sup_{h \in \HH}\abs{\widehat{\rdv{C}}_{n}(h) - \crit_{\ddist}(h)}.
\end{align}
Assuming $\crit_{\ddist}$ takes its minimum on $\HH$, an obvious reference point $h^{\ast} \in \HH$ is a minimizer of the expected test loss, i.e., $\crit_{\ddist}(h^{\ast}) = \crit_{\ddist}^{\ast} \defeq \inf\{\crit_{\ddist}(h): h \in \HH\}$. This $\crit_{\ddist}^{\ast}$ represents the best off-sample performance we can hope to achieve, and thus it is natural to ask that the learning algorithm be able to get arbitrarily close to this benchmark with sufficiently high probability, given enough data. More precisely, we can make the following requirement:
\begin{align}\label{eqn:perf_consistency}
\forall\,\varepsilon > 0, \qquad \lim\limits_{n \to \infty} \prr\left\{ \crit_{\ddist}(\rdv{H}_{n}) - \crit_{\ddist}^{\ast} > \varepsilon \right\} = 0.
\end{align}
This is just convergence of $\crit_{\ddist}(\rdv{H}_{n}) \to \crit_{\ddist}^{\ast}$ in probability, also known as \term{consistency} of the learning algorithm. There is no conceptual reason to limit this analytical framework to the zero-one loss; \emph{any} random loss $\loss(h) \defeq \ell(h;\rdv{Z})$ with finite expectation $\crit_{\ddist}(h) = \exx_{\ddist}\loss(h)$ can in principle be considered. Minimizing the expected value of a general-purpose loss over a hypothesis class $\HH$ is the ``general setting'' of learning formulated by \citet[Ch.~1]{vapnik1999NSLT}.

The requirement (\ref{eqn:perf_consistency}) can be stated in a different but equivalent fashion as follows: for any desired accuracy $\varepsilon > 0$ and confidence $0 < \delta < 1$, there must exist $N(\varepsilon,\delta) \in \NN$ such that for any $n \geq N(\varepsilon,\delta)$, we have
\begin{align}\label{eqn:perf_PAC}
\prr\left\{ \crit_{\ddist}(\rdv{H}_{n}) - \crit_{\ddist}^{\ast} > \varepsilon \right\} \leq \delta.
\end{align}
The condition (\ref{eqn:perf_PAC}) emphasizes \emph{precision} and \emph{confidence} requirements explicitly through the parameters $\varepsilon$ and $\delta$ respectively. The highly influential work of \citet{valiant1984a} considers a learning model which emphasizes precisely these points, with a particular interest in the computational complexity of achieving the $(\varepsilon,\delta)$-condition in (\ref{eqn:perf_PAC}). This became known as the \term{PAC-learning} model.\footnote{The original idea of \citep{valiant1984a} was fleshed out further in subsequent works by \citet{valiant1984b,valiant1985a}. The naming of this model given in the main text (PAC: probably approximately correct) appears to date back to \citet{angluin1988a} and \citet{angluin1988b}. The textbook of \citet{kearns1994CLTIntro} provides good additional background.} In effect, the work of Valiant brought the consistency requirement to the community of computer scientists working on machine learning.\footnote{While the PAC-learning model was the first to receive widespread attention, clearly the general setting of Vapnik supersedes the PAC-learning model; the $(\varepsilon,\delta)$-condition (\ref{eqn:perf_PAC}) appears from the start of a book by \citet[Sec.~1.6]{vapnik1982EDBED}, a 1982 English translation of a Russian book published in 1979. However, the work of Vapnik was not well-known by computer scientists in the West until the late 1980s, in particular due to the influential work of \citet{blumer1989a}, who introduced and built upon the seminal paper of \citet{vapnik1971a}. As an aside, according to \citet[Ch.~2]{vapnik2006Afterword}, L.~Valiant did not know of his work when authoring his influential 1984 article \citep{valiant1984a}. On the other hand, many probability theorists in the West were familiar with the work of Vapnik and Chervonenkis; see for example \citet{dudley1978a} or \citet{pollard1984ConvSP}.} In this community, as is mentioned by \citet{kearns1994a}, it should be noted that the emphasis on the expected value of a generic loss (i.e., the general setting of Vapnik) was largely due to the learning model studied by \citet{haussler1992a}, in which the classical risk $\exx_{\ddist}\loss(h)$ plays a central role.

To conclude this section, given the historical context established in the preceding paragraphs, we give a brief overview highlighting how ``learning as expected loss minimization'' is emphasized in many influential and widely-read textbooks. \citet{kearns1994CLTIntro} give an elementary introduction to the PAC-learning framework and basics of VC theory, with emphasis placed on computational tractability and tasks of interest in computer science. It goes without saying that all the books of Vapnik \citep{vapnik1982EDBED,vapnik1998SLT,vapnik1999NSLT} are chiefly concerned with the problem of expected loss minimization. While restricted to the pattern recognition problem (binary classification), the highly influential text of \citet{devroye1996ProbPR} characterizes the learning problem in terms of $\exx_{\ddist}\left[\loss(\rdv{H}_{n}) \cond \Z_{n} \right]$, formulated using the consistency condition (\ref{eqn:perf_consistency}). Expected loss is also central to the learning setup in the textbook of \citet{ripley1996PRNN}. This setup is carried on by \citet{anthony1999NNTheory}, in their text on neural network learning. A more modern textbook dealing with statistical and computational learning theory is \citet{mohri2012Foundations}, and the expected loss plays a foundational role. Expected loss is used as a central criterion in the introductory text of \citet{shalev2014a}, which also emphasizes links between stochastic optimization and machine learning. It is clear that these theory-oriented textbooks are centered around the expected loss, but this notion also appears frequently in more applied machine learning textbooks aimed at introducing key methods and tools to a general audience, albeit in more of a supplementary role. To give a few examples, we have \citet[Ch.~36]{mackay2003InfoTheoryLearning}, \citet[Ch.~1]{bishop2006PRML}, \citet[Ch.~5--6]{murphy2012MLPP}, \citet[Ch.~5]{goodfellow2016DeepLearning}, among others. Given this bibliographic evidence, along with the historical context provided in section \ref{sec:history_expectation} earlier, it is evident that the notion of expected loss as a metric for off-sample generalization has been pervasive in both the theory and practice of modern machine learning.

\section{A Look Forward}

In the preceding \S{\ref{sec:intro}}--\S{\ref{sec:history}}, it has been our aim to provide the reader with a unified introductory look at the key classes of criteria that have been studied and applied in the machine learning literature, with the traditional mean-driven paradigm providing important historical context for recent developments. Within this paper, the base loss function $\ell(\cdot)$, the decision set $\HH$, and the desirability set $\CC$ were all left abstract, with focus placed on the learning criterion $\crit_{\ddist}(\loss)$. Moving forward, a natural direction of interest is to ask how these four elements can interact with each other. Consider the following broad points of inquiry:
\begin{itemize}
\item Starting with $\ell$ and $\crit_{\ddist}$, when can we reduce the problem (\ref{eqn:new_pac_condition}) (under $\ell$) into problem (\ref{eqn:classic_pac_condition}) under a different loss function?

\item Can we identify conditions (on $\ell$, $\HH$, and $\ddist$) under which traditional norm-based regularization can be emulated using an appropriately designed $\crit_{\ddist}$?

\item How does the theory of surrogate function design change under intractable loss functions (e.g., zero-one) when $\crit_{\ddist}$ does not depend linearly on $\ddist$? (e.g., $\crit_{\ddist}(\loss) = \vaa_{\ddist}\loss$)
\end{itemize}
Beyond these points of technical interest, there is the glaring methodological problem that in many cases, it simply is not clear which criterion is ``best'' for a given problem, since this is often subjective, with precise numerical thresholds difficult to pin down, and will differ from case to case. Solutions to this problem (e.g., metric elicitation \citep{ali2022arxiv}) are still in their infancy, but there is no question that the development of principled, transparent, and accessible procedures for criteria design, ideally tied in with the back-end algorithmic details, is an important frontier for machine learning research.

\section*{Acknowledgments}

This work was supported by JST ACT-X Grant Number JPMJAX200O and JST PRESTO Grant Number JPMJPR21C6. We also would like to thank Professor Takashi Washio for constructive feedback on an early version of this paper.

\bibliography{./refs/refs}

\end{document}